\documentclass[screen,nonacm]{acmart}
\usepackage{colortbl}
\usepackage[utf8]{inputenc}
\usepackage[framemethod=default]{mdframed} 
\usepackage{tikz} 
\usepackage{enumitem}
\usepackage{adjustbox} 
\usepackage{array} 
\usepackage{amsbsy} 
\usepackage{multicol} 
\usepackage[longtable]{multirow}
\usepackage{caption} 
\usepackage{subcaption} 
\usepackage{setspace} 
\newsavebox{\measurebox} 

\usepackage{afterpage} 
\usepackage{longtable} 

\mdfsetup{skipabove=5pt,skipbelow=5pt}
\mdfdefinestyle{examplestyle}{
linecolor=gray,
linewidth=0.75pt,
frametitlerule=true,
frametitlebackgroundcolor=gray!10,
userdefinedwidth=0.95\linewidth,
align=center,
frametitleaboveskip=1pt,
frametitlebelowskip=1pt
}
\mdtheorem[style=examplestyle]{example}{Example}
\mdtheorem[style=examplestyle]{definition}{Definition}
\mdtheorem[style=examplestyle]{inclusion_criterion}{Inclusion Criterion}
\mdfdefinestyle{keyideastyle}{
linecolor=gray,
linewidth=1.5pt,
leftmargin=1cm,
rightmargin=1cm,
innertopmargin=0.75pt,
innerbottommargin=0.75pt,
skipabove=0.1pt,
skipbelow=0pt}

\definecolor{verylightgray}{gray}{0.97}

\newcolumntype{R}[2]{%
    >{\adjustbox{angle=#1,lap=\width-(#2)}\bgroup}%
    l%
    <{\egroup}}
\newcommand*\rot{\multicolumn{1}{R{90}{1em}}}

\usepackage{pdfrender}
\newcommand*{\boldcheckmark}{%
  \textpdfrender{
    TextRenderingMode=FillStroke,
    LineWidth=.8pt, 
  }{\checkmark}%
}

\newcolumntype{g}{>{\columncolor{verylightgray}}c}

\usepackage{array,booktabs}
\newcommand {\otoprule}{\midrule [\heavyrulewidth]}
\newcolumntype {+}{ >{\global\let\currentrowstyle\relax}}
\newcolumntype {^}{ >{\currentrowstyle }}
  \newcommand {\rowstyle}[1]{\gdef\currentrowstyle{#1} %
  #1\ignorespaces
  }
\newcommand{\tabhead}{\rowstyle{\bfseries}}

\renewcommand{\arraystretch}{1.1} 

\setcopyright{rightsretained} 
\begin{document}

\title[A Systematic Review on Evaluating Explainable AI]{From Anecdotal Evidence to Quantitative Evaluation Methods: A Systematic Review on Evaluating Explainable AI}

\author{Meike Nauta}
\email{m.nauta@utwente.nl}
\orcid{0000-0002-0558-3810}
\affiliation{%
  \institution{University of Twente}
  \city{Enschede}
  \country{the Netherlands}
}
\affiliation{%
  \institution{University of Duisburg-Essen}
  \city{Essen}
  \country{Germany}
}
\author{Jan Trienes}
\email{jan.trienes@uni-due.de}
\orcid{0000-0001-8891-0022}
\affiliation{%
  \institution{University of Duisburg-Essen}
  \city{Essen}
  \country{Germany}
}
\author{Shreyasi Pathak}
\email{s.pathak@utwente.nl}
\orcid{0000-0002-6984-8208}
\affiliation{%
  \institution{University of Twente}
  \city{Enschede}
  \country{the Netherlands}
}
\affiliation{%
  \institution{University of Duisburg-Essen}
  \city{Essen}
  \country{Germany}
}
\author{Elisa Nguyen}
\email{t.q.e.nguyen@alumnus.utwente.nl}
\orcid{0000-0003-0224-268X}
\author{Michelle Peters}
\email{xai@michellepeters.eu}
\orcid{0000-0001-8884-730X}
\affiliation{%
  \institution{University of Twente}
  \city{Enschede}
  \country{the Netherlands}
}

\author{Yasmin Schmitt}
\orcid{0000-0002-1948-7475}
\author{J{\"o}rg Schl{\"o}tterer}
\email{joerg.schloetterer@uni-due.de}
\orcid{0000-0002-3678-0390}
\affiliation{%
  \institution{University of Duisburg-Essen}
  \city{Essen}
  \country{Germany}
}

\author{Maurice van Keulen}
\email{m.vankeulen@utwente.nl}
\orcid{0000-0003-2436-1372}
\affiliation{%
  \institution{University of Twente}
  \city{Enschede}
  \country{the Netherlands}
}

\author{Christin Seifert}
\email{christin.seifert@uni-due.de}
\orcid{0000-0002-6776-3868}
\affiliation{%
  \institution{University of Duisburg-Essen}
  \city{Essen}
  \country{Germany}
}

\renewcommand{\shortauthors}{Nauta et al.}

\acmJournal{CSUR}
\acmYear{2023} \acmVolume{1} \acmNumber{1} \acmArticle{1} \acmMonth{1} \acmPrice{}\acmDOI{10.1145/3583558}

\begin{abstract}
The rising popularity of explainable artificial intelligence (XAI) to understand high-performing black boxes raised the question of how to evaluate explanations of machine learning (ML) models. 
While interpretability and explainability are often presented as a subjectively validated binary property, we consider it a multi-faceted concept. We identify 12 conceptual properties, such as Compactness and Correctness, that should be evaluated for comprehensively assessing the quality of an explanation. Our so-called Co-12 properties serve as categorization scheme for systematically reviewing the evaluation practices of more than 300 papers published in the last 7 years at major AI and ML conferences that introduce an XAI method. We find that 1 in 3 papers evaluate exclusively with anecdotal evidence, and 1 in 5 papers evaluate with users. This survey also contributes to the call for objective, quantifiable evaluation methods by presenting an extensive overview of quantitative XAI evaluation methods. Our systematic collection of evaluation methods provides researchers and practitioners with concrete tools to thoroughly validate, benchmark and compare new and existing XAI methods. The Co-12 categorization scheme and our identified evaluation methods open up opportunities to include quantitative metrics as optimization criteria during model training in order to optimize for accuracy and interpretability simultaneously. 
\end{abstract}



\maketitle
\clearpage
\section{Introduction}
The last decades have seen rapid development and extensive usage of Artificial Intelligence (AI) and Machine Learning~(ML). The size and complexity of these models grew in pursuit of predictive performance. However, the focus on accuracy alone is increasingly coming under criticism,
since it leaves us with big black-box models with non-transparent decision making which prevents users from assessing, understanding and potentially correcting the system. The necessity for interpretable and explainable AI (XAI) therefore arises, aiming to make AI systems and their results more understandable to humans~\cite{adadi_peeking_2018}. Especially the emergence of deep learning in the last decade has led to a high interest in developing methods for explaining and interpreting black-box systems.

With an increasing number of XAI methods, the demand grows for suitable XAI evaluation metrics~\cite{barredo_arrieta_explainable_2020,leavitt_towards_2020,adadi_peeking_2018,guidotti2018survey,burkart2021survey}. This need is not only recognized by the AI community; also the Human Computing Interaction (HCI) community is concerned with developing transferable evaluation methods for XAI~\cite{upol_chi_2021}. In addition, a research agenda for Hybrid Intelligence~\cite{research_agenda_hybrid_akata} has explicitly formulated a research question asking how the quality and strength of explanations can be evaluated.
Whereas traditional performance indicators exist to evaluate prediction accuracy and computational complexity, auxiliary criteria such as interpretability may not be easily quantified~\cite{doshi-velez_considerations_2018}. This difficulty is part of the reason for the huge variation in explanation techniques, and the optimal evaluation methods and measures could depend on the application domain, the type of explanation, the type of data, the background knowledge of the user and the question to be answered. The XAI community has yet to agree upon standardized evaluation metrics to go beyond often reported anecdotal evidence showing individual, convincing examples that pass the first test of having ``face-validity''~\cite{doshi-velez_considerations_2018}. Evaluation is then only based on ``the researchers' intuition of what constitutes a good explanation''~\cite{miller_explanation_2017}. 
The lack of quantitative evaluation impedes interpretability research, since anecdotal inspection is not sufficient for robust verification~\cite{leavitt_towards_2020}. Many authors~(e.g.~\cite{adebayo_sanity_2018,leavitt_towards_2020,lage_evaluation_2019,Jacovi_towards_2020}) argue that relying on such anecdotal evidence alone is insufficient and that other aspects of the explanations should be evaluated as well.  

Whereas interpretability can be presented as a binary property, we consider interpretability a multi-faceted characteristic and argue that a quantitative way of measuring interpretability should result in a multi-dimensional view indicating the extent to which certain properties are satisfied. Having a set of quantitative, and preferably automated, metrics for various properties would allow researchers and practitioners to a) assess and validate the interpretability of a single explanation method and its explanations, b) objectively compare and benchmark multiple explanation methods, c) add interpretability as optimization criteria during model training to tune the accuracy-interpretability trade-off.

\paragraph{Contributions.} 
This survey contributes to the demand for XAI evaluation methods with a systematic review on the evaluation of explainability and interpretability\footnote{Regarding terminology: `interpretability' and `explainability' are closely related and often used interchangeably in the XAI context~\cite{carvalho_machine_2019,burkart2021survey}. We equate them in this survey as well. The same holds for `explainable artificial intelligence' and `interpretable machine learning'. See also Section~\ref{sec:definitions_terminology}.} methods.  
Specifically, we collected 606 papers (2014-2020) published at twelve flagship computer science conferences in a structured manner, of which 312 introduced an XAI method~(Section~\ref{sec:paper_selection}). Our categorization of explainable AI methods is available on our interactive website at \url{https://utwente-dmb.github.io/xai-papers/}. Analysis of this set of papers provide quantitative insights into the extent and nature of research activity in XAI and the evaluation of the resulting explanations (Section~\ref{sec:results_g2_g3}). For instance, we have found that feature importance is the most common explanation type and that the majority of XAI methods explain single predictions rather than providing global insights about the model reasoning. Moreover, 1 in 3 papers evaluate exclusively with anecdotal evidence, and 1 in 5 papers evaluate with a user study. 
Additionally, we argue that explainability is a multi-faceted concept and make this explicit with our \emph{Co-12 properties} of explanation quality. We use Co-12 as categorization scheme for the analysis of quantitative evaluation methods in papers that introduce, apply or evaluate an XAI method (361 papers in total). As a result, Section~\ref{sec:quantitative_evaluation_metrics} presents an overview of quantitative evaluation and benchmarking methods for explainable AI. Hence, we address the frequently reported lack of quantitative evaluation methods~\cite{gilpin_explaining_2018,adadi_peeking_2018,barredo_arrieta_explainable_2020,guidotti2018survey,leavitt_towards_2020} and respond to the call for automated and quantifiable evaluation metrics for robust and falsifiable explainability research~\cite{leavitt_towards_2020}. We hope that our collection of evaluation methods will facilitate a more complete and inclusive evaluation for objectively validating and comparing new and existing XAI methods. Our overview can serve as a handbook for researchers and practitioners that are looking for suitable evaluation methods to evaluate multiple aspects of their XAI method.
Lastly, Section~\ref{sec:research_opportunities} discusses the implications of our results and identifies research opportunities for the XAI domain. The most promising potential we see for XAI research is to shift from evaluating explanations to incorporating explanation quality metrics into the training process to optimize for explainability.

\paragraph{Reader's Guide.}
Since Explainable AI is of interest to a broad range of people from XAI researchers to ML practitioners, we provide guidance for various types of readers as follows:
\begin{itemize}
    \item \textbf{All readers:} Table~\ref{tab:desiderata_overview} presents our Co-12 properties, a high-level decomposition of explanation quality, such as Completeness, Correctness and Compactness. Our definition of \emph{explanation} is introduced in Section~\ref{sec:definitions_terminology}.
    \item \textbf{Readers who want to \emph{evaluate} XAI methods:} Table~\ref{tab:evaluation_methods_descriptions_papers} summarizes our collection of automated, quantitative evaluation methods. This overview provides tools for researchers and practitioners to thoroughly validate and compare XAI methods. Section~\ref{sec:quantitative_evaluation_metrics} describes each automated evaluation method and its variations in more detail. 
    \item \textbf{Readers interested in \emph{trends} and \emph{evaluation practices} in XAI:} Sections~\ref{sec:results_g1}~and~\ref{sec:results_g2_g3} quantitatively summarize our findings on research activity in XAI (2014-2020) and evaluation practices: from anecdotal evidence to user studies. E.g., we find that 1 in 5 papers evaluate with users.
    Additionally, Section~\ref{sec:research_opportunities} provides a unifying, conclusive view on XAI evaluation practices and presents research opportunities. 
    \item \textbf{Readers \emph{looking} for available XAI methods:} Our website \url{https://utwente-dmb.github.io/xai-papers/} with 312 papers that introduce an explainable AI method, is a useful starting point for anyone searching for specific explainable AI methods. Based on our categorization as presented in Figure~\ref{fig:review_protocol_categorisation}, readers can filter XAI methods on e.g.~\emph{type of data}, \emph{type of explanation} and \emph{type of task}. Different explanation types are summarized in Table~\ref{tab:method:explanation-types-overview}. 
    \item \textbf{Readers interested in theory of XAI evaluation:} Section~\ref{sec:related_work} is recommended as background reading, as it summarizes related work, presents pros and cons of evaluating with users and discusses the discrepancy between objective and subjective evaluation.
\end{itemize}

\paragraph{Comparison with other surveys.} 
In contrast to most XAI surveys that review explainability methods, we focus on the \emph{evaluation} of explainability.
Some surveys discuss evaluation as part of a broader review of XAI methods~\cite{adadi_peeking_2018,gilpin_explaining_2018, montavon_methods_2018,zhang_visual_2018,mohseni_multidisciplinary_2019,nunes2017systematic,carvalho_machine_2019,burkart2021survey,das2020opportunities,joshi_review_xai_2021, chen_iml_mythos_diagnostics}, or mainly discuss evaluation with user studies~\cite{lage_evaluation_2019, hoffman_metrics_2019,chromik_taxonomy_2020,mohseni_multidisciplinary_2019}. Others discuss XAI evaluation within a limited scope, by focusing on an application domain or subarea of XAI: e.g.~on the healthcare domain~\cite{markus_role_2021}, the evaluation of local explanations~\cite{guidotti_evaluating_local_2021}, contrastive and counterfactual explanations~\cite{stepin2021survey} or explanations and their evaluation for recommender systems~\cite{Tintarev2015_explaining,nunes2017systematic}. Closest related to our work is the survey of Zhou et al.~\cite{zhou_evaluating_2021} that presents a concise overview and discussion regarding evaluating XAI without a structured literature review, and the work of Vilone and Longo~\cite{vilone2021notions} that present a list of concepts related to explainability and discuss evaluation methods in combination with explainable AI methods. 
In contrast to previous surveys, we conducted a large-scale, \emph{systematic} review on the \emph{evaluation} of explainability in a broad context, resulting in qualitative and quantitative insights and therefore aiming to offer guidance on future XAI evaluations. 

\subsection{Definitions and Terminology}
\label{sec:definitions_terminology}
Explanations have been discussed for decades in many research areas. However, various contexts may require different types of explanations. No definition therefore precisely captures the scope of all different settings. 
In this survey, we focus on the context of \emph{explainable artificial intelligence} and \emph{interpretable machine learning}. Those terms, together with \emph{explainability} and \emph{interpretability} (and to some extent also \emph{intelligibility}) are often used interchangeably~\cite{carvalho_machine_2019,burkart2021survey,murdoch_interpretable_2019,acl/HaseB20,molnar2020interpretable}. Some argue that the terms are closely related but distinguish between them~\cite{montavon_methods_2018,barredo_arrieta_explainable_2020,gilpin_explaining_2018,roscher_2020_explainable}, although there is no consensus on what the distinction exactly is~\cite{Bellucci_terminology_2021}. We equate them (and use them interchangeably) to keep a general, inclusive discussion and to ensure that we do not exclude work because of different terminology. We consider related concepts such as fairness, safety, causality, ethical decision making and privacy~\cite{barredo_arrieta_explainable_2020,lipton_mythos_2018,doshi-velez_considerations_2018} out of scope.

We frame explanations in the context of \emph{explainable artificial intelligence} and define an explanation as follows:
\begin{definition*}
An \textbf{explanation} is a presentation of (aspects of) the reasoning, functioning and/or behavior of a machine learning model in human-understandable terms.
\label{defintion:explanation}
\end{definition*}

The definition of explanation is inspired by work of van Lent et al.~\cite{vanlent2004explainable} who coined the term XAI. Quoting van Lent et al.~\cite{vanlent2004explainable}: ``Ideally, this Explainable AI can present the user with an easily understood chain of reasoning from the user’s order, through the AI’s knowledge and inference, to the resulting behavior''. The fact that the explanation should be ``easily understood'' is also emphasized by others: ``systems are interpretable if their operations can be understood by a human''~\cite{biran2017explanation} and ``an interpretation is the mapping of an abstract concept into a domain that the human can make sense of''~\cite{montavon_methods_2018}. We adopt the phrasing of Doshi-Velez and Kim~\cite{doshi-velez_considerations_2018} who define interpretability as the ability ``to explain or to present in understandable terms to a human''.

We specifically included ``reasoning, functioning and/or behavior'' in our definition to capture different types of explanations, which can be roughly related to the three approaches identified by Gilpin et al.~\cite{gilpin_explaining_2018}. \emph{Reasoning} refers to the process on how a model came to a particular decision. In Gilpin's terms, it explains the ``processing'' of data to answer the question ``Why does this particular input lead to that particular output?''. \emph{Functioning} refers to the (internal) workings and internal data structures of the machine learning models, and therefore relates to the ``representation of data''~\cite{gilpin_explaining_2018}. \emph{Behavior} refers to how the model globally operates without specifically analyzing the internal workings (e.g. by observing input and output) which can ``simplify interpretation''~\cite{gilpin_explaining_2018}. The inclusive \emph{or} indicates that an explanation can satisfy multiple goals. 

\paragraph{Terminology and Notation} In this paragraph, we introduce our terminology with corresponding notation to allow an unambiguous review and discussion in the rest of this paper.
Let $f$ be a \emph{predictive machine learning model}, such as a neural network or decision tree, trained to take some data as input and predict the corresponding output. Given input $x$, the predictive model $f$ outputs a prediction $f(x)$. Let $e$ denote an \emph{explanation method} which generates explanations. The explanation method can produce a local explanation which explains a single prediction, denoted as $e(f(x))$, or a global explanation explaining the predictive model as a whole: $e(f)$. The explanations generated by explanation method $e$ are of a certain \emph{explanation type}, e.g.~a decision tree, heatmap or rule list. 
In case of an intrinsically interpretable model (also called \emph{self-explaining}~\cite{nips/Alvarez-MelisJ18}), the predictive model is already explainable by design, since interpretability is built into the architecture. For such self-explaining methods, the model and explanation method are the same, i.e., $e=f$, such that $f(x) = e(x)$. 

\section{Related Work: Anecdotal Evidence, Functional Evaluation and User Studies}
\label{sec:related_work}
We identified two main themes in the literature on XAI evaluations:  1) the difference between evaluating plausibility and correctness of an explanation, 2) XAI evaluation with or without users. The following two subsections summarize opinions from existing literature on each of these themes.

\subsection{Evaluating Plausibility or  Correctness of an Explanation}
\label{sec:plausibility_vs_correctness}
Whereas standard evaluation metrics exist to evaluate the performance of a predictive model, there is no agreed-upon evaluation strategy for explainable AI. As a result, a common evaluation strategy is to show individual, potentially cherry-picked, examples that look reasonable~\cite{murdoch_interpretable_2019} and pass the first test of having ``face-validity''~\cite{doshi-velez_considerations_2018}. 
Many authors argue that relying on such anecdotal evidence alone is insufficient and can even be ``misleading''~\cite{adebayo_sanity_2018}. 
Leavitt and Morcos~\cite{leavitt_towards_2020} note that researchers too frequently assume that an explanation method and the resulting explanation are faithful. ``Intuition is essential for building understanding'' but ``unverified intuition [...] can facilitate misapprehension''~\cite{leavitt_towards_2020}. They argue that the lack of quantitative evaluation impedes interpretability research, since anecdotal inspection is not sufficient for robust verification. 

Related, several papers warn that evaluating the plausibility and convincingness of an explanation to humans is different from evaluating its correctness, and these evaluation criteria should not be conflated. Jacovi and Goldberg~\cite{Jacovi_towards_2020} argue that it is not guaranteed that a plausible explanation is also truthfully reflecting the reasoning of the model. Petsiuk et al.~\cite{petsiuk_rise_2018} believe that ``keeping humans out of the loop for evaluation makes it more fair and true to the classifier’s own view on the problem rather than representing a human’s view''. Gilpin et al.~\cite{gilpin_explaining_2018} explain that an unreasonable-looking explanation could indicate either an error in the reasoning of the predictive model, or an error in the explanation producing method. Visual inspection on the \emph{plausibility} of the explanation, such as anecdotal evidence, cannot make this distinction. We can relate this to the well-known phrase ``garbage in, garbage out": when the machine learning model is trained on flawed data, it learns nonsensical relations which are in turn shown by the explanation. The explanation might then be perceived as being wrong, although it is truthfully reflecting the model's reasoning. Zhang et al.~\cite{zhang_towards_2019} identify this as main shortcoming when evaluating explainable AI and state that checking whether an explanation ``looks reasonable'' only evaluates the accuracy of the black box model and is not evaluating the faithfulness of the explanation. Adebayo et al.~\cite{adebayo_sanity_2018} motivate this issue with a clear example: they show that saliency maps to explain computer vision tasks can be highly similar to edge detectors. Visual inspection would be insufficient to differentiate edge detection from model-sensitive explanations. ``Here the human observer is at risk of confirmation bias when interpreting the highlighted edges as an explanation of the class prediction.''~\cite{adebayo_sanity_2018}. 

These commentaries relate to the inherent coupling of evaluating the black box' predictive accuracy with explanation quality. As pointed out by Robnik-Šikonja and Bohanec~\cite{robnik_perturbation-based_2018}, the correctness of an explanation and the accuracy of the predictive model may be orthogonal. Although the correctness of the explanation is independent of the correctness of the prediction, visual inspection cannot distinguish between the two. Samek et al.~\cite{samek_evaluating_2017} analyze this issue for heatmaps that explain computer vision algorithms: ``the heatmap quality does not only depend on the algorithms used to compute a heatmap, but also on the performance of the classifier, whose efficiency largely depends on the model being used, and the amount and quality of available training data.''. 
Gilpin et al.~\cite{gilpin_explaining_2018} find it unethical to optimize an explanation towards hiding undesirable attributes. They argue that explanation methods should be evaluated on ``how they behave on the curve from maximum interpretability to maximum completeness''~\cite{gilpin_explaining_2018}.

\subsection{Evaluating With or Without User Studies}
\label{sec:functional_vs_human}
Related to the discussion on evaluating plausibility is the discussion on evaluating with or without users.
Doshi-Velez and Kim~\cite{doshi-velez_considerations_2018} propose to categorize the evaluation of interpretability with a 3-level taxonomy. The top level contains \emph{application-grounded evaluation} and involves human subject experiments with domain experts within a real application, such that the method can be evaluated by the intended users with respect to a particular task. The second level contains \emph{human-grounded evaluation} and involves user studies with lay persons on simplified tasks that ``maintain the essence of the target application''~\cite{doshi-velez_considerations_2018}. This evaluation level is suitable when researchers want to evaluate more general notions of explanation quality instead of one particular end-goal, or when reaching the target user is difficult due to e.g.~high costs or a low number of available domain experts. The controlled human experiments can result in subjective results by asking users for perceived quality, or objective results by measuring performance of participants on specific tasks. 

The third level of the taxonomy~\cite{doshi-velez_considerations_2018} contains the \emph{functionally-grounded evaluation} approach, which includes evaluation where human experiments are not needed but instead uses computational proxy measures for interpretability. For example, measuring the size of the explanation or validating feature importance by perturbing model input.
Doshi-Velez and Kim~\cite{doshi-velez_considerations_2018} discuss the potential advantages of this evaluation type: besides saving time and costs and therefore being more scalable, it can be particularly appropriate when user studies are unethical or when the method is not yet mature enough for evaluation with users. However, they also emphasize that these proxy metrics are best suited once user studies have already confirmed the interpretability of the model class. 
This is in accordance with Miller et al.~\cite{miller_explainable_2017} who state that proxy metrics are valid evaluations, but the authors support more intensive human evaluations to have ``real-world impact''.  

Others put extra arguments forward in favor of automated metrics where no user involvement is needed. User studies for machine learning research often depend on online crowd platforms such as Amazon Mechanical Turk which can lead to ethical issues. Critics argue that these platforms are largely unregulated and that workers are poorly compensated~\cite{hara2018data,semuels_2018_poorlypaid}. Results from user studies are also ``rarely replicable or even comparable''~\cite{cvpr/WangV20}.
Additionally, besides saving time and resources~\cite{markus_role_2021}, Herman~\cite{herman_promise_2017} and Ancona et al.~\cite{ancona_unied_2017} argue that user studies imply a strong bias towards simpler explanations that are closer to the user's expectations, ``at the cost of penalizing those methods that might more closely reflect the network behavior''~\cite{ancona_unied_2017}. "A good explanation method should not reflect what humans attend to, but what task methods attend to"~\cite{acl/SchutzeRP18}. 
This relates to the discussion in Section~\ref{sec:plausibility_vs_correctness} on anecdotal evidence as evaluation strategy. Validating explanations with users can unintentionally combine the evaluation of explanation correctness with evaluating the correctness of the predictive model. 
Leavitt and Morcos therefore plead for ``clear, specific, testable and falsifiable hypotheses'' that dissociate the evaluation of the explanation method from the predictive model~\cite{leavitt_towards_2020}. Quantitative evaluation also allows a formal comparison between various explanation methods~\cite{markus_role_2021}, and contrasting them under different applications and purposes~\cite{barredo_arrieta_explainable_2020}. In the context of usability evaluation in the HCI community, Greenberg and Buxton~\cite{greenberg_2008_usability} argue that there is a risk of executing user studies in an early design phase, since this can quash creative ideas or promote poor ideas. Miller therefore argues that proxy studies are especially valid in early development~\cite{miller_explainable_2017}. Qi et al.~\cite{qi_embedding_2021} indicate that ``evaluating explanations objectively without a human study is also important because simple parameter variations can easily generate thousands of different explanations, vastly outpacing the speed of human studies''. 

\subsection{Discussion: Relation to Our Work}
The discussions in the field have illustrated that evaluation of explanations is not self-evident and shows various pitfalls. We also found that explainability is indeed not a binary property, and that various aspects of an explanation should be evaluated independently of each other. Since knowledge about which aspects constitute a good explanation is scattered, this survey presents an aggregated view of \emph{what to evaluate} by introducing twelve properties on explanation quality (Section~\ref{sec:evaluation_practice}). Our property `Correctness' addresses the faithfulness of an explanation with respect to the predictive model, whereas `Coherence' addresses the plausibility of an explanation. Hence, both properties can be evaluated separately. Additionally, Section~\ref{sec:quantitative_evaluation_metrics} presents an overview of quantitative evaluation methods and categorizes them along our so-called Co-12 properties. This contribution concretely addresses \emph{how} to evaluate different aspects of an explanation, and therefore provides conceptual guidance to the XAI community. We specifically focus on functionally-grounded evaluation. A concise overview of evaluation methods with user studies is provided in Supplementary Material.

\section{Methodology}
\label{sec:methodology}
\subsection{Paper Selection}
\label{sec:paper_selection}

\begin{figure}[t!]
    \centering
    \includegraphics[width=\linewidth]{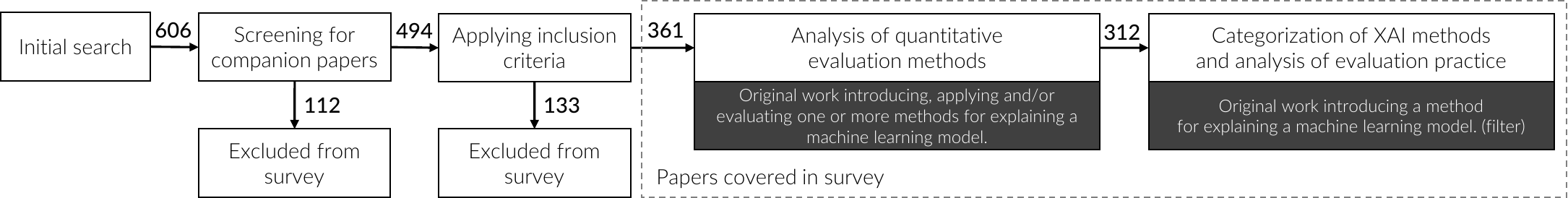}
    \caption{Flow diagram showing the number of papers through the different phases of the reviewing process. A more detailed description of our paper collection process, inclusion and exclusion criteria, and reviewing process is available in Supplementary Material. }
    \label{fig:inclusion_flowdiagram}
\end{figure}

We collected papers in a structured manner to provide both quantitative and qualitative insights about the XAI domain based on a large corpus of scientific work on XAI evaluation methods. 
Since the literature on XAI is highly diverse and distributed across different (sub)disciplines, we selected literature published from 2014 to 2020 at one of the following twelve prominent conferences: AAAI, IJCAI, NeurIPS (formerly NIPS), ICML, ICLR, CVPR, ICCV, ACL, WWW, ICDM, SIGKDD (also called KDD), SIGIR. We used DBLP\footnote{\url{https://dblp.org/}} to conduct a keyword search in publication titles with the following search query: \texttt{explain* OR explanat* OR interpret*} to capture terms including \emph{explainable}, \emph{explaining}, \emph{explanation}, \emph{interpretable} and \emph{interpretability}.
This search, conducted on 4th of May 2021, resulted in 606 papers. We manually excluded companion papers (such as workshop papers and tutorials), resulting in 494 papers. To only include relevant papers in our analysis, we apply the following \textbf{inclusion criterion}:

\noindent \emph{Original work introducing, applying and/or evaluating one or more methods for explaining a machine learning model}.

Applying the inclusion criteria led to 361 papers being included, as shown in Figure~\ref{fig:inclusion_flowdiagram}. Subsequently, we can apply a \textbf{filter} that only selects the papers that \emph{introduce} an XAI method, resulting in 312 papers. We apply this filter to analyze how introduced XAI methods are evaluated when they are first presented (Section~\ref{sec:results_g2_g3}). For collecting all evaluation methods, we review all 361 included papers since 49 papers do not introduce a new XAI method, but could contain relevant evaluation metrics to compare and evaluate existing XAI methods. We do not want such papers to skew our quantitative results in Section~\ref{sec:results_g2_g3}, but include them in our evaluation overview in Section~\ref{sec:quantitative_evaluation_metrics} for completeness. A more detailed description of our paper collection process, inclusion and exclusion criteria, and reviewing process is available in Supplementary Material. 

\subsection{Review Protocol: Categorization of Explainable AI Methods}
\label{sec:xai_categorisation}
\begin{figure}[tbh]
    \centering
    \includegraphics[width=0.85\linewidth]{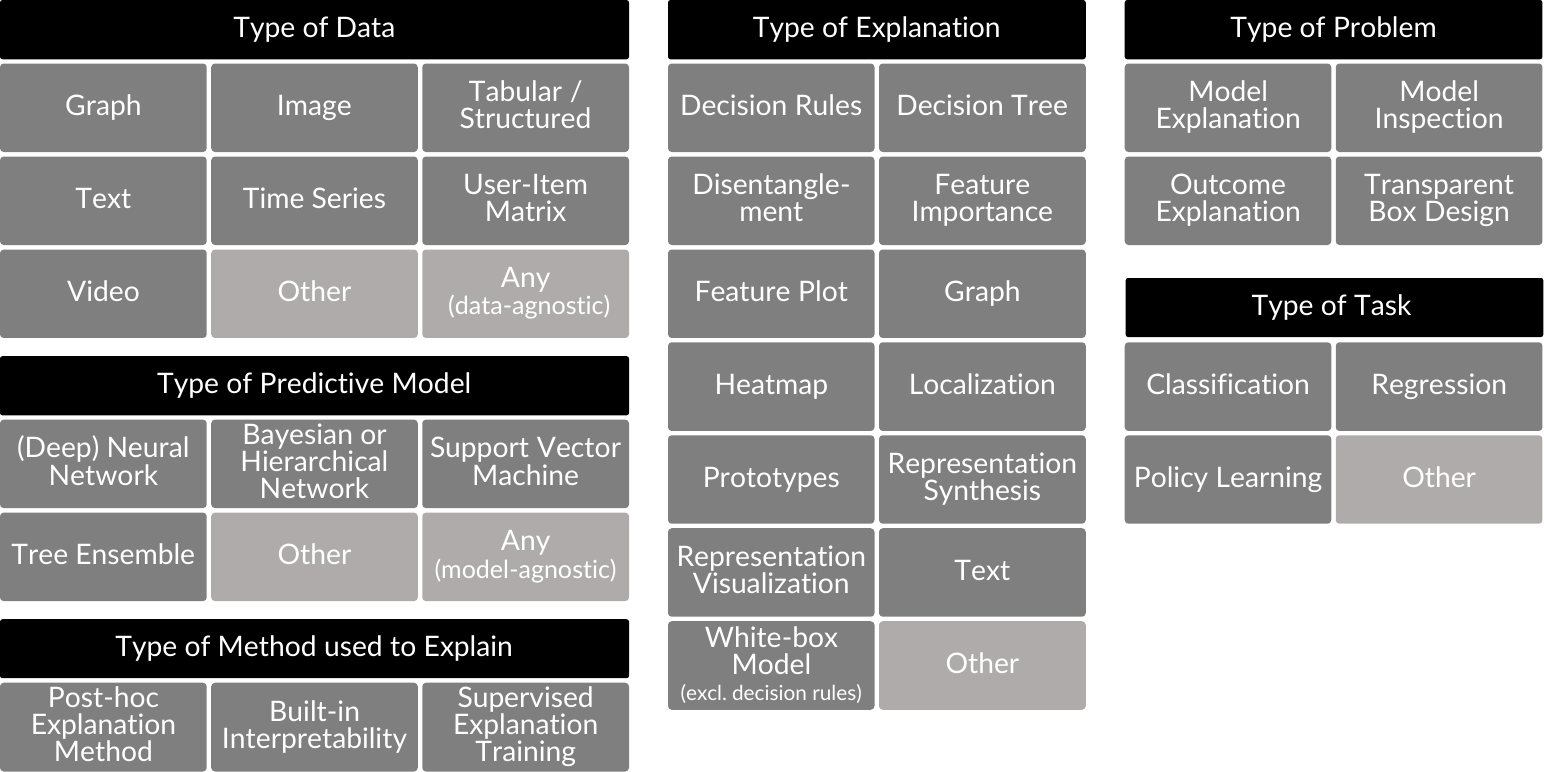}
    \caption{Categorization of explainable AI methods along 6 dimensions: type of data used as input $x$ to predictive model $f$, type of explanation, type of explanation problem that is addressed, type of predictive model to be explained ($f$), type of task for which model $f$ is used, and the type of method used to explain. Papers can address multiple categories per dimension.}
    \label{fig:review_protocol_categorisation}
\end{figure}

For further analysis of the included papers, we categorize each paper and analyze the properties of the explanation method and the evaluation of the explanations in more detail.
For each included paper, we review the main content and do not consider appendices and supplementary material. Each included paper is first categorized along six dimensions, 
in order to create a structured overview of XAI methods. 
Figure~\ref{fig:review_protocol_categorisation} summarizes the 6 dimensions and their corresponding categories. The following paragraphs discuss three dimensions in more detail. 

We adopt the taxonomy of Guidotti et al.~\cite{guidotti2018survey}, presenting four \textbf{types of problems} that an XAI method can solve: (i) Model Explanation --  globally explaining model $f$ through an interpretable, predictive model; (ii) Model Inspection -- globally explaining some specific property of  model $f$ or  its prediction;
(iii) Outcome Explanation -- explaining an outcome/prediction of $f$ on a particular input instance; (iv) Transparent Box Design -- the explanation method is an interpretable model (i.e., $e=f$) also making the predictions.
Note that a set of outcome explanations can collectively comprise a global explanation for model inspection (cf. e.g.~\cite{nips/FryeRF20, aaai/Ribeiro0G18, nips/LuoCXYZC020}). 
For details regarding this taxonomy, we refer to Section 4 of the survey by Guidotti et al.~\cite{guidotti2018survey}.

We identified three different types of general \textbf{types of methods} used to explain a machine learning model: i) Post-hoc explanation methods (also called \emph{reverse engineering}~\cite{guidotti2018survey}): explain an already trained predictive model;  ii) Interpretability built into the predictive model, such as white-box models, attention mechanisms or interpretability constraints (e.g. sparsity) included in the training process of the predictive model; and iii) Supervised explanation training, where a ground-truth explanation is provided in order to train the model to output an explanation.

Evaluation methods are often specific for a specific type of explanation, thus we categorize each paper by \textbf{type of explanation}. In contrast to most XAI surveys, we disregard the explanation construction approach but focus on the explanation's output format. 
Since we could not find a complete and recent overview of explanation types, we reviewed recent XAI surveys~\cite{das2020opportunities, molnar2020interpretable,arya2019one,xie2020explainable} and adapted and extended the categories identified by Guidotti et al.~\cite{guidotti2018survey} with these insights. We grouped some explanation types and separated others based on the expected difference in evaluation metrics resulting in the 14 categories outlined in Table~\ref{tab:method:explanation-types-overview}.
Note that some explanation methods combine multiple explanation types, thus the categories are not mutually exclusive w.r.t. explanation methods. We do not consider counterfactual explanations a separate category. Although counterfactual explanations answer a different type of question, the type of explanation utilized is still one of the above categories. 

\begin{table}[tb]
\caption{Overview of types of explanations. }
\label{tab:method:explanation-types-overview}
\begin{tabular}{+p{0.15\linewidth}^p{0.8\linewidth}}
\toprule
\tabhead
Category & Description and Examples\\\otoprule
    Decision Rules    & Logical rules, incl. decision sets~\cite{kdd/LakkarajuBL16}, anchors~\cite{aaai/Ribeiro0G18}, decision tables~\cite{huysmans_empirical_2011} and programs~\cite{icml/VedantamDLRBP19}. \\[3pt]
    Decision Tree      & Rooted graph with conditional statement at each node, e.g.~ProtoTree~\cite{nauta2020neural}.\\[3pt]
    Disentanglement    & Disentangled representation, where each disjoint feature might have a semantic meaning, e.g.~InfoGAN~\cite{nips/ChenCDHSSA16}.
    \\[3pt]
   Feature \newline Importance & Set of 1-dimensional non-binary values/scores to indicate feature relevance, feature contribution or attribution. A feature is not necessarily an input feature to predictive model $f$, but it should be a feature in the explanation. Examples include SHAP~\cite{nips/LundbergL17} and importance scores by LIME~\cite{kdd/Ribeiro0G16}.
    \\[3pt]
    Feature Plot    & Plot or figure showing relations or interactions between features or between feature(s) and outcome. Examples include Partial Dependence Plot~\cite{friedman_pdp_2001}, Individual Conditional Expectation plot~\cite{goldstein_ice_2015} and Feature Auditing~\cite{adler2018auditing}.
    \\[3pt]
    Graph & Graphical network structure with nodes and edges, e.g.~Abstract Policy Graph~\cite{aaai/TopinV19}, Knowledge graph~\cite{aaai/WangWX00C19}, Flow graph~\cite{aaai/RustamovK18} and Finite State Automata~\cite{hou_automata_2020}.
    \\[3pt]
    Heatmap &  Map with at least 2 dimensions visually highlighting non-binary feature attribution, activation, sensitivity, attention or saliency. Includes attention maps~\cite{iccv/SelvarajuCDVPB17}, perturbation masks~\cite{iccv/FongV17} and Layer-Wise Relevance Propagation~\cite{bach_lrp_2015}.
    \\[3pt]
    Localization & Binary feature importance. Features can be any type of covariate used in the explanation, such as words, tabular features, or bounding boxes. Examples include binary maps with image patches~\cite{kdd/Ribeiro0G16}, segmentation~\cite{nips/HoyerMKKF19} and bounding boxes~\cite{aaai/ZhangCWZ17}.
    \\[3pt]
    Prototypes & (Parts of) Representative examples, including concepts~\cite{icml/KimWGCWVS18}, influential training instances~\cite{acl/HanWT20}, prototypical parts~\cite{nips/ChenLTBRS19,nauta2020neural}, nearest neighbors and criticisms~\cite{nips/KimKK16}.
    \\[3pt]
    Representation Synthesis & Artificially produced visualization to explain representations of the predictive model. Examples include generated data samples~\cite{iclr/SinglaPCB20}, Activation Maximization~\cite{nguyen_synthesizing_2016} and feature visualization~\cite{olah2017feature}.
    \\[3pt]
    Representation Visualization & Charts or plots to visualize representations of the predictive model, including visualizations of dimensionality reduction with scatter plots~\cite{vandermaaten_tsne_2008}, visual cluster analysis~\cite{icdm/LiuMWH0G20} and Principal Component Analysis.
    \\[3pt]
    Text & Textual explanation via natural language, e.g.~\cite{acl/RajaniMXS19, nips/CamburuRLB18}.
    \\[3pt]
    White-box Model & Intrinsically interpretable models. Predictive model $f$ is interpretable and therefore acts as explanation. Examples include a scoring sheet~\cite{berk_rudin_risk_2017} and linear regression. Decision Rules and Decision Trees do not fall into this category, since they are categories on their own. 
    \\[3pt]
    Other & Explanation that does not fit any other category.
    \\\bottomrule
\end{tabular}
\end{table}

\subsection{Review Protocol: Evaluation of XAI methods with Co-12 Properties}
\label{sec:evaluation_practice}

\begin{figure}[t!]
    \centering
    \includegraphics[width=0.7\linewidth]{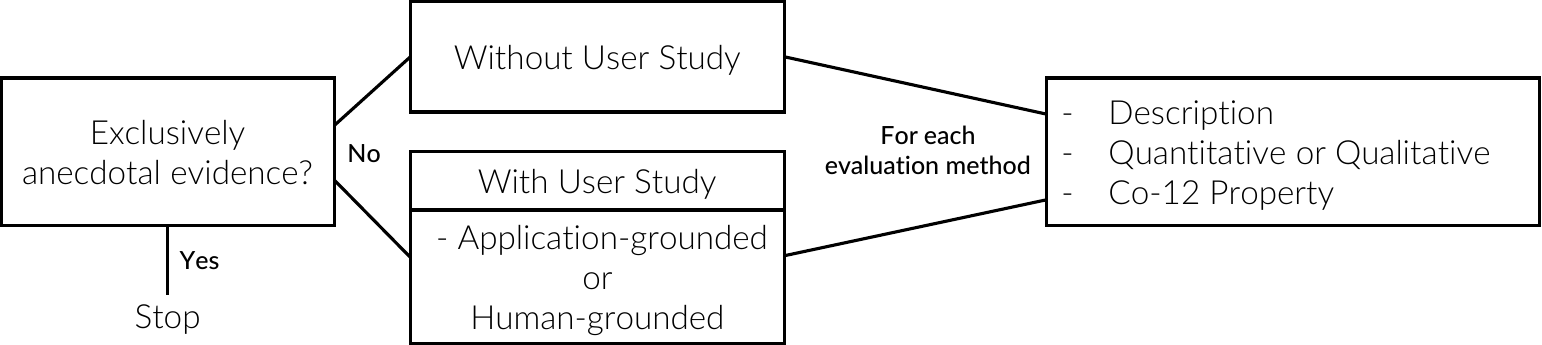}
    \caption{Our review protocol when reviewing the evaluation of an XAI method.}
    \label{fig:review_protocol_structure}
\end{figure}

When reviewing the evaluation of an XAI method, we distinguish between evaluation with users and without users. We focus on the evaluation of explanation method $e$ and/or its produced explanations. Hence, this is not about the evaluation of predictive model $f$, and we do not take the evaluation of predictions into account. Therefore, evaluation metrics that evaluate $f$ or the predictions of $f$, such as task accuracy and computation time, are not included. Additionally, evaluation metrics for $e$ that do not \emph{directly} influence explanation quality (such as run time or construction overhead of $e$~\cite{swartout_explanation_1993}), are also excluded. Lastly, we excluded quantitative methods that use explanations to get more insights into the predictive model. Such approaches are not about \emph{evaluating} explanations, but rather \emph{utilize} explanations to investigate model $f$. For example, explainable AI is used to analyze whether model $f$ is fair~\cite{sigir/FuXGZHGXGSZM20} or biased~\cite{iccv/MicheliniLLJ19,iclr/SinglaPCB20}, whether $f$ is right for the right reasons~\cite{icml/RiegerSMY20}, or whether $f$ is overfitting~\cite{aaai/ChenJ20}. 

As summarized in Figure~\ref{fig:review_protocol_structure}, we analyzed for each paper whether exclusively anecdotal evidence is presented for evaluating the quality of the XAI method. For all other evaluation methods, we collected a short description on how it works, whether the measure is qualitative or quantitative and which property is evaluated. Additionally, for user studies, we assess whether the study is application-grounded, i.e. in a real application with domain experts (following the taxonomy of~\cite{doshi-velez_considerations_2018}) or human-grounded, i.e. using simplified, but similar tasks with lay persons.

\textbf{Co-12 Explanation Quality Properties.}
Different aspects regarding explanation quality can be evaluated, as also discussed in Section~\ref{sec:plausibility_vs_correctness}. We therefore argue that explainability is a non-binary characteristic, that can be measured by evaluating to what degree certain properties are satisfied. Based on conceptual literature that discusses explanation quality and properties of a good explanation, we identified twelve desired explanation properties that together present an aggregated view of \emph{what to evaluate}. We paid specific attention to covering as much of the reviewed properties as possible, minimizing semantic overlap between properties and grouping different terminology that describe a similar property. 
Our so-called \emph{Co-12 properties} (pronounce as co-twelve) regarding explanation quality are presented below and summarized in Table~\ref{tab:desiderata_overview}. 

\begin{table}[t!]
  \centering
    \caption{Our Co-12 explanation quality properties, grouped by their most prominent dimension: Content, Presentation or User.}
  \label{tab:desiderata_overview}
  \begin{tabular}{p{0.015\linewidth}p{0.21\linewidth} p{0.72\linewidth}}
  \toprule
     & Co-12 Property & Description\\
    \midrule
    \multirow{12}{1.5cm}{\rotatebox[origin=c]{90}{Content}}&\textbf{Correctness} & Describes how faithful the explanation is w.r.t. the black box.\\
    & & \cellcolor{gray!25}\textbf{Key idea:} Nothing but the truth\\
    & \textbf{Completeness} & Describes how much of the black box behavior is described in the explanation.\\
    & & \cellcolor{gray!25}\textbf{Key idea:} The whole truth\\
    & \textbf{Consistency} & Describes how deterministic and implementation-invariant the explanation method is.\\
    & & \cellcolor{gray!25}\textbf{Key idea:} Identical inputs should have identical explanations\\
    &\textbf{Continuity} & Describes how continuous and generalizable the explanation function is.\\
    & & \cellcolor{gray!25}\textbf{Key idea:} Similar inputs should have similar explanations\\
    &\textbf{Contrastivity} & Describes how discriminative the explanation is w.r.t. other events or targets.\\
    & & \cellcolor{gray!25}\textbf{Key idea:} Answers ``why not?'' or ``what if?'' questions\\
    & \textbf{Covariate complexity} & Describes how complex the (interactions of) features in the explanation are.\\
    & & \cellcolor{gray!25}\textbf{Key idea:} Human-understandable concepts in the explanation \\
    \midrule
    \multirow{6}{1.5cm}{\rotatebox[origin=c]{90}{Presentation}}&\textbf{Compactness} & Describes the size of the explanation. \\
    & & \cellcolor{gray!25}\textbf{Key idea:} Less is more\\
    &\textbf{Composition} & Describes the presentation format and organization of the explanation. \\
    & & \cellcolor{gray!25}\textbf{Key idea:} \emph{How} something is explained\\
    &\textbf{Confidence} & Describes the presence and accuracy of probability information in the explanation.\\
    & & \cellcolor{gray!25}\textbf{Key idea:} Confidence measure of the explanation or model output\\
    \midrule
    \multirow{6}{1.5cm}{\rotatebox[origin=c]{90}{User}}& \textbf{Context} & Describes how relevant the explanation is to the user and their needs.\\
    & & \cellcolor{gray!25}\textbf{Key idea:} How much does the explanation matter in practice?\\
    & \textbf{Coherence} & Describes how accordant the explanation is with prior knowledge and beliefs.\\
    & & \cellcolor{gray!25}\textbf{Key idea:} Plausibility or reasonableness to users\\
    &\textbf{Controllability} & Describes how interactive or controllable an explanation is for a user.\\
    & & \cellcolor{gray!25}\textbf{Key idea:} Can the user influence the explanation?\\
     \bottomrule
  \end{tabular}

\end{table}

\textbf{Correctness} addresses the truthfulness/faithfulness of the explanation with respect to predictive model $f$, the model to be explained. Hence, it indicates how truthful the explanations are compared to the ``true'' black box behavior (either locally or globally). Note that this property is not about the predictive accuracy of the black box model, but about the \emph{descriptive accuracy} of the explanation~\cite{murdoch_interpretable_2019}. Ideally, an explanation is ``nothing but the truth''~\cite{kulesza_too_2013}, and high correctness is desired~\cite{ijcai/BhattWM20, kulesza_too_2013, nips/Alvarez-MelisJ18, atanasova-etal-2020-diagnostic, Jacovi_towards_2020,robnik_perturbation-based_2018, sokol_explainability_2020,swartout_explanation_1993, wiltschko_evaluating_2020, yang_evaluating_2019, kulesza_principles_2015, zhang_towards_2019}. 

\textbf{Completeness} addresses the extent to which the explanation explains predictive model $f$. Ideally, the explanation provides ``the whole truth''~\cite{kulesza_too_2013}. High completeness is desired~\cite{kulesza_principles_2015,zhou_evaluating_2021, sokol_explainability_2020, zhang_towards_2019, cui_integrative_2019, silva_towards_2018, zhou_evaluating_2021} in order to provide enough detail, but it should be balanced with compactness and correctness: ``don't overwhelm''~\cite{kulesza_principles_2015}. 
    \begin{itemize}[noitemsep,topsep=0pt]
        \item \textbf{Reasoning-completeness} indicates the extent to which the explanation describes the entire internal dynamic of the model~\cite{zhou_evaluating_2021}. One extreme is ``revealing all the mathematical operations and parameters in the system''~\cite{gilpin_explaining_2018} such as white-box models which are by definition fully reasoning-complete. The other extreme are global surrogate models that are trained to give the same predictions as black box $f$, without considering any internal reasoning of $f$. A design choice should be made regarding the reasoning-completeness by selecting an explanation type suited for a specific context. Therefore, reasoning-completeness is often only evaluated qualitatively to compare different explanation types. 
        \item \textbf{Output-completeness} addresses the extent to which the explanation covers the output of model $f$. Thus, it is a ``quantification of unexplainable feature components''~\cite{zhang_towards_2019} and measures how well the explanation method agrees with the predictions of the original predictive model~\cite{carvalho_machine_2019,Jacovi_towards_2020}. 
    \end{itemize}

\textbf{Consistency} checks that identical inputs have identical explanations~\cite{honegger_shedding_2018, andrews_survey_1995}. In practice, this property addresses to what extent the explanation method is deterministic. Additionally, for explanation methods that do not consider the internals of the black box but only observe input and output, consistency regards \emph{implementation invariance} which states that two models that give the same outputs for all inputs should have the same explanations~\cite{carvalho_machine_2019,robnik_perturbation-based_2018}. Atanasova et al.~\cite{atanasova-etal-2020-diagnostic} add that models with the same architecture but trained from different random seeds should give the same explanations when they follow the same reasoning path. For explanation methods that do consider the internals of the black box, Montavon~\cite{montavon_gradient-based_2019} argues that implementation invariance is still a desired property, but should then be evaluated without changing the actual function. 

\textbf{Continuity} considers how continuous (i.e. smooth) the explanation function is that is learned by the explanation method. A continuous function ensures that small variations in the input, for which the model response is nearly identical, do not lead to large changes in the explanation~\cite{ijcai/BhattWM20,carvalho_machine_2019,yang_evaluating_2019, honegger_shedding_2018, robnik_perturbation-based_2018, nips/Alvarez-MelisJ18,wiltschko_evaluating_2020, montavon_gradient-based_2019, atanasova-etal-2020-diagnostic}. Continuity also adds to generalizability beyond a particular input~\cite{miller_explanation_2017, sokol_explainability_2020} or generalizability to new contexts~\cite{srinivasan_explanation_2020}.

\textbf{Contrastivity} addresses the discriminativeness of an explanation and aims to facilitate comparisons in relation to other targets or events~\cite{carvalho_machine_2019}. Miller argues that an explanation should not only explain an event, but explain it ``\emph{relative to some other event} that did not occur''~\cite{miller_explanation_2017}. Honegger~\cite{honegger_shedding_2018} adds the separability property that non-identical instances from different populations must have dissimilar explanations.

\textbf{Covariate complexity} considers the complexity of the covariates (i.e. features) used in the explanation in terms of semantic meaning and interactions between the covariates and the target. The covariates in the explanation should be comprehensible~\cite{carvalho_machine_2019}, and ``concepts should have an immediate human-understandable interpretation''~\cite{nips/Alvarez-MelisJ18}. This could mean that the variables used in the explanation are different from the features given as input to model $f$, since ``interpretable data representations'' are desired~\cite{kdd/Ribeiro0G16}.
Also non-complex interactions between features are desired, such as monotonicity~\cite{doshi-velez_considerations_2018}. Wilson et al.~\cite{wilson_2015_human_kernel} found that humans favor smooth and simpler functions, and have inductive biases towards recognizable patterns, such as step functions or a sawtooth pattern.

\textbf{Compactness} considers the size of the explanation and is motivated by human cognitive capacity limitations. Explanations should be sparse, short and not redundant to avoid presenting an explanation that is too big to understand~\cite{ijcai/BhattWM20,zhou_evaluating_2021,miller_explanation_2017, silva_towards_2018, cui_integrative_2019, sokol_explainability_2020}.

\textbf{Composition} considers the presentation format, organization and structure of the explanation~\cite{carvalho_machine_2019}, such that the way in which the explanation is presented to the user increases its ``clarity''~\cite{zhou_evaluating_2021}. As mentioned by Huysmans et al.~\cite{huysmans_empirical_2011}, ``some representation formats are generally considered to be more easily interpretable than others''. Hence, this property is about \emph{how} something is explained instead of \emph{what} is explained. Examples include the usage of higher-level information~\cite{nips/Alvarez-MelisJ18}, abstractions~\cite{doshi-velez_considerations_2018, swartout_explanation_1993} or suitable terminology~\cite{swartout_explanation_1993}, and not using explanations that are circular~\cite{srinivasan_explanation_2020}. Others compare the interpretability of different representation formats, such as Booth et al.~\cite{ijcai/BoothMS19} evaluating logical sentences of different forms to investigate how to
best present propositional theories to humans, and Huysmans et al.~\cite{huysmans_empirical_2011} comparing the comprehensibility of decision tables, trees and rule-based models.  

\textbf{Confidence} concerns whether the explanation has a measure of certainty or other probability information. It can reflect two facets of certainty: i) a confidence measure of the black box prediction~\cite{carvalho_machine_2019, robnik_perturbation-based_2018,yang_evaluating_2019,atanasova-etal-2020-diagnostic}, or ii) the truthfulness or likelihood of the explanation~\cite{miller_explanation_2017, silva_towards_2018, atanasova-etal-2020-diagnostic}. Opinions are divided about the last facet of this property, since it is argued that referring to probabilities might not be so effective, since people have difficulties to correctly estimate probabilities~\cite{miller_explanation_2017}.

\textbf{Context} addresses the extent to which the user and their needs are taken into account for comprehensible explanations~\cite{carvalho_machine_2019}. Explanations should be relevant to the user's needs and level of expertise~\cite{miller_explanation_2017, sokol_explainability_2020}. Srinivasan and Chander~\cite{srinivasan_explanation_2020} argue, from a cognitive science perspective, that explanations should not only serve AI scientists, but a whole variety of stakeholders, e.g.~policy makers and customers.

\textbf{Coherence} assesses to what extent the explanation is consistent with relevant background knowledge, beliefs and general consensus~\cite{carvalho_machine_2019, miller_explanation_2017, sokol_explainability_2020} and hence addresses reasonableness~\cite{gilpin_explaining_2018}, plausibility~\cite{Jacovi_towards_2020} and ``agreement with human rationales''~\cite{atanasova-etal-2020-diagnostic}. It is often argued that evaluating coherence alone (with e.g.~anecdotal evidence) is not sufficient~\cite{leavitt_towards_2020}, and that coherence and correctness should not be conflated but evaluated separately~\cite{Jacovi_towards_2020}. 
Note that this property addresses \emph{external} coherence, and is different from \emph{internal} coherence to indicate that parts in an explanation fit together~\cite{zemla_evaluating_2017}. 

\textbf{Controllability} indicates to what extent a user can control, correct or interact with an explanation~\cite{carvalho_machine_2019,miller_explanation_2017,kulesza_principles_2015, sokol_explainability_2020}, since it is argued that ``explanations are social''~\cite{miller_explanation_2017}.

\section{Overall Statistics of Included Papers}
\label{sec:results_g1}

\begin{figure}[tbh]
\centering
\begin{subfigure}[b]{.52\textwidth}
\centering
  \includegraphics[width=0.97\linewidth]{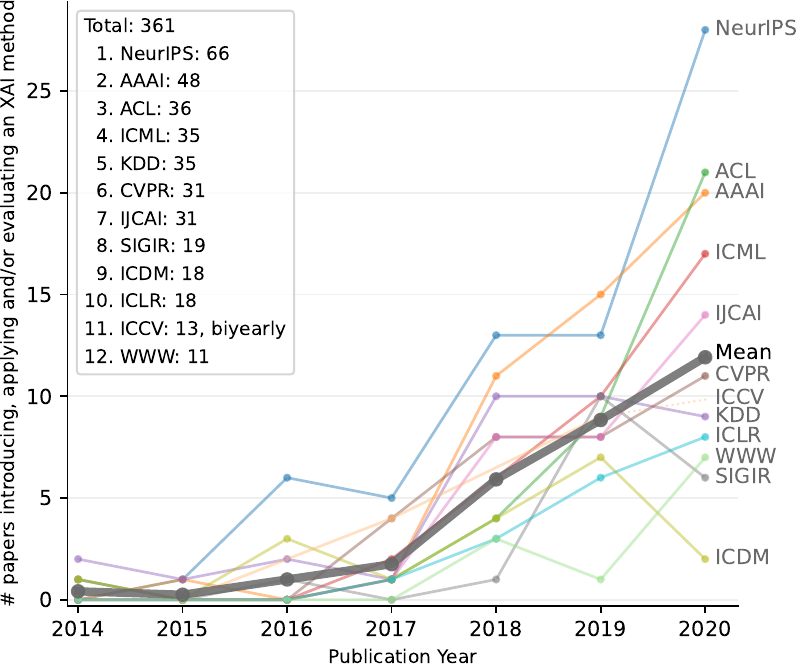}
  \caption{Number of included papers per venue that introduce, apply or evaluate an XAI method (gray line shows average). The box on the left lists the total number of included papers.}
  \label{fig:linegraph_papers_venue_year}
\end{subfigure}%
\hfill
\begin{subfigure}[b]{.45\textwidth}
\centering
  \includegraphics[width=0.9\linewidth]{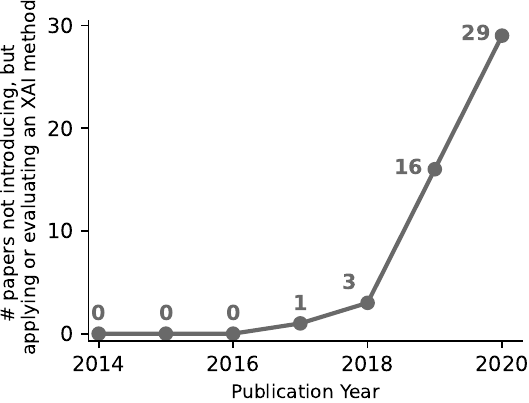}
    \caption{Number of included papers that do not introduce, but rather apply and/or evaluate an XAI method, plotted by publication year.}
    \label{fig:linegraph_only_G4}
\end{subfigure}
\caption{Number of included papers per publication year. }
\label{fig:two_linefigures}
\end{figure}

Figure~\ref{fig:linegraph_papers_venue_year} shows that the number of papers on explainable AI and interpretable machine learning is growing over the years. There were few papers in our set published in 2014 or 2015 (5 and 3 in total, respectively), which could be related to the fact that the topic itself was less popular or that terms as `interpretability' and `explainable AI' were not yet used in these years~\cite{rudin2019stop}.
We see a steady increase of included papers since 2016. Especially 2018 shows a significant increase, which corresponds with findings by Barredo Arrieta et al.~\cite{barredo_arrieta_explainable_2020}. However, our paper selection does not show an exponential growth as found by Adadi and Berrada~\cite{adadi_peeking_2018}, although some conferences such as ACL and ICML do show an exponential increase. NeurIPS (formerly NIPS) is in our dataset the conference with the most papers on explainable AI and interpretable ML. Especially the large jumps in 2018 and 2020 are striking. Also at AAAI the number of papers on explainability increased substantially over the last years. 

Additionally, we analyzed the number of papers that do not introduce an XAI method, but apply or evaluate them (i.e., the 49 papers that would be excluded by the filter as shown in Figure~\ref{fig:inclusion_flowdiagram}). Figure~\ref{fig:linegraph_only_G4} shows that this number increased substantially in 2019 and again in 2020. This trend could indicate that the awareness regarding evaluation and comparison of XAI methods has grown in the last years, which again could point towards an increasing maturity of the field. 

\section{Statistics on XAI Methods and their Evaluation}
\label{sec:results_g2_g3}
\begin{figure}[t!]
\centering
    \includegraphics[width=\linewidth]{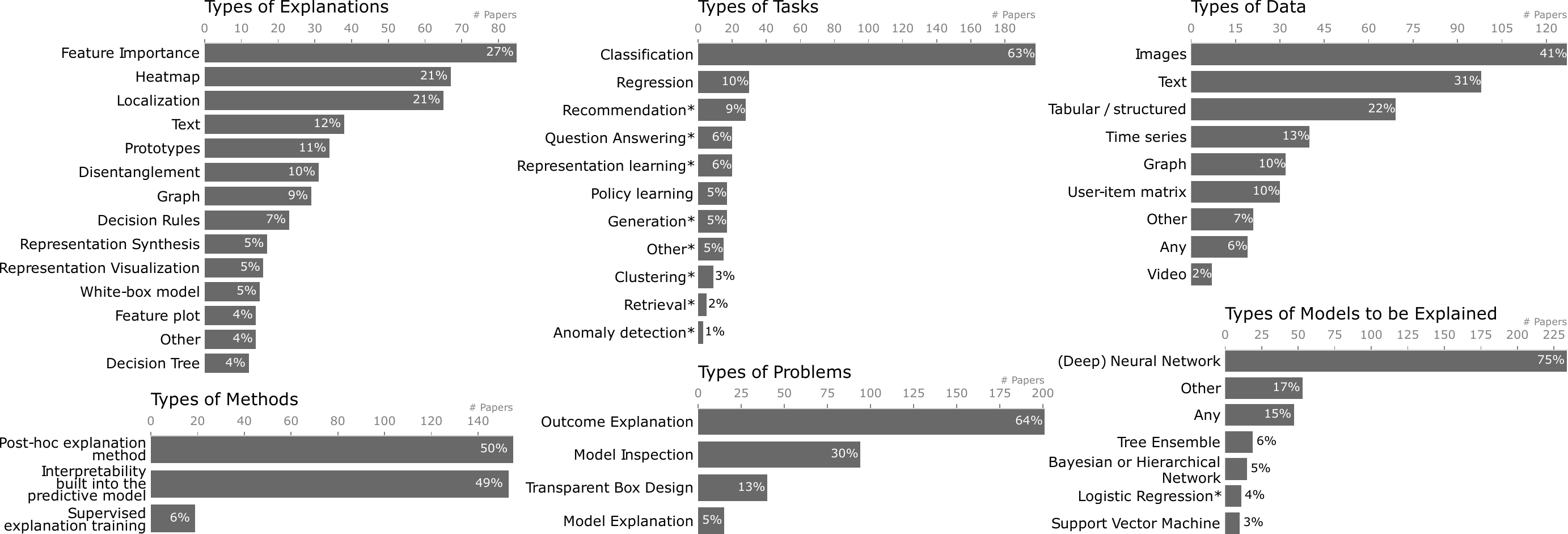}
    \caption{Categorization of papers introducing an explainable AI method, following the six dimensions as presented in Section~\ref{sec:xai_categorisation}. Note that categories are non-exclusive, so a paper can fall into multiple categories per dimension. *: category is manually added after the reviewing process and might therefore not be complete (i.e. high precision, potentially low recall).}
    \label{fig:barcharts_all_G2}
\end{figure}

In this section we analyze the 312 papers that introduce a method for explaining a machine learning model. An interactive website for this labeled dataset is available at \url{https://utwente-dmb.github.io/xai-papers/}. Figure~\ref{fig:barcharts_all_G2} presents summary statistics regarding the categorization of XAI methods. Supplementary material presents a more detailed analysis of the categorization of XAI methods. In the remainder of this section, we focus on the \emph{evaluation} of XAI methods. 
\newpage
\textbf{Our summary statistics on XAI evaluation:}
\vspace{-\topsep}
\begin{itemize}[noitemsep]
    \item \textbf{33\%} only evaluated with anecdotal evidence
    \item \textbf{58\%} applied quantitative evaluation
    \item \textbf{22\%} evaluated with human subjects in a user study, of which 23\% evaluated with domain experts, i.e. application-grounded~\cite{doshi-velez_considerations_2018}.
\end{itemize}

\begin{figure}[t!]
    \centering
    \begin{subfigure}[b]{0.48\linewidth}
    \centering
        \includegraphics[width=\linewidth]{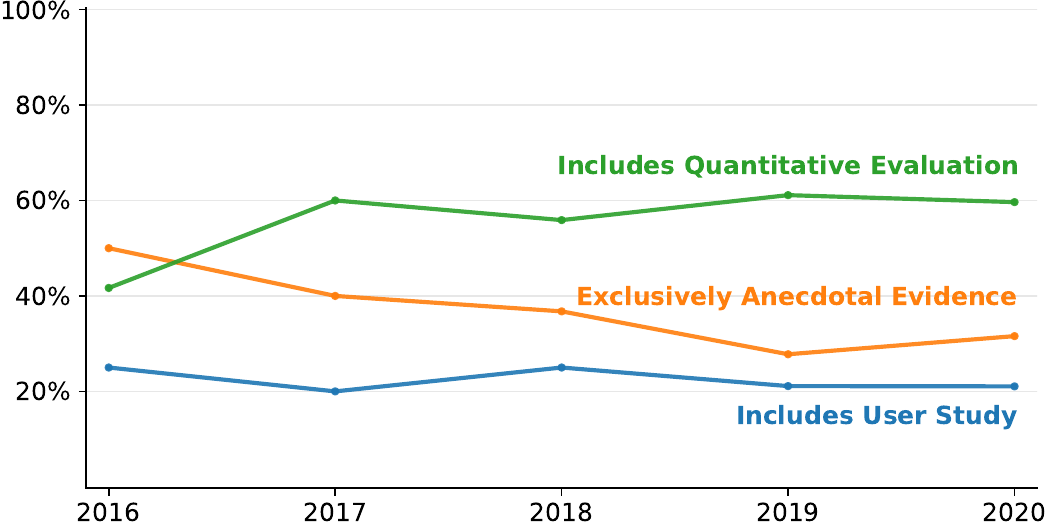}
    \caption{Evaluation practices of the 312 papers that introduce a method for explaining a machine learning model.}
    \label{fig:linegraph_evaluation_G3}
    \end{subfigure}\quad
    \begin{subfigure}[b]{0.48\linewidth}
    \centering
    \includegraphics[width=0.88\linewidth]{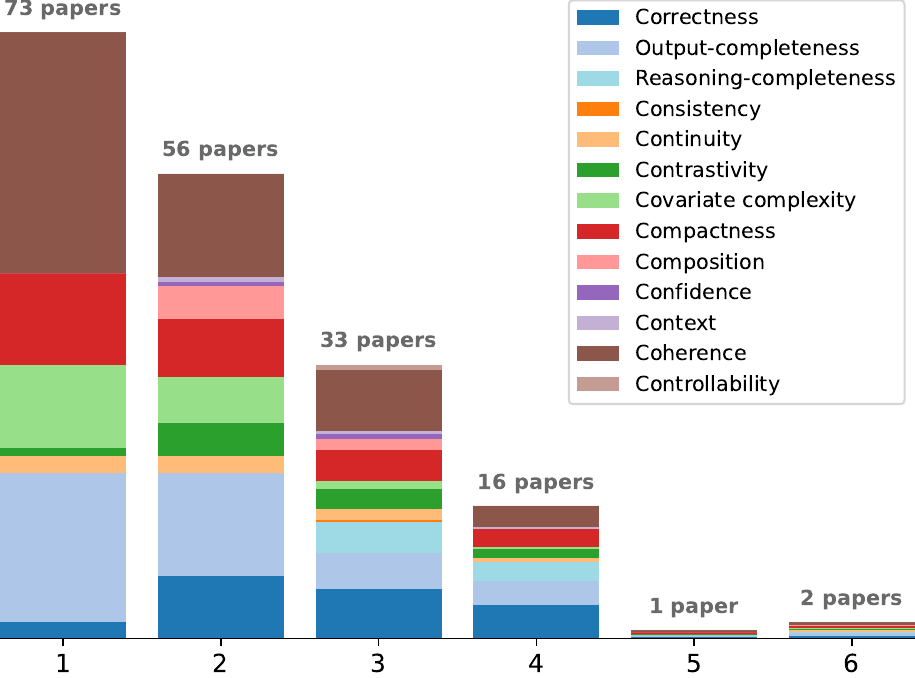}
    \caption{Total number of unique Co-12 properties quantitatively evaluated in a paper that introduces an XAI method.}
    \label{fig:stacked_barchart_co12}
    \end{subfigure}
    \caption{XAI evaluation practice.}
\end{figure}

Earlier research has found that few papers quantitatively evaluated their explanations. Only 5\% of the papers analyzed by Adadi et al.~(2018)~\cite{adadi_peeking_2018} evaluated their interpretable machine learning method and quantified its relevance. Nunes and Jannach reported that only 21\% of their 190 analyzed studies (1990-2017) that presented an XAI technique or tool contained ``any form of evaluation, except from toy examples''~\cite{nunes2017systematic}. We have reviewed more papers, including more recent papers, to shed light on the evaluation practices from 2014 to 2020. Our statistics regarding the usage of quantitative evaluation are higher, which, although possibly influenced by the venues we collected from, indicates that the evaluation of XAI has become more extensive over the years. 
Some of the papers that do not quantitatively evaluate their XAI method argue that their model architecture is inherently interpretable, and therefore do not explicitly evaluate its interpretability. Lipton~\cite{lipton_mythos_2018} however notes that even white-box models might not be interpretable anymore when their size exceed the limited capacity of human cognition, and Jacovi and Goldberg~\cite{Jacovi_towards_2020} suggest that intrinsically interpretable methods should be held to the same standards as post-hoc interpretation methods with similar evaluation methods. Others do not explicitly evaluate their method with quantitative evaluation metrics, but present mathematical theory to support their claims~(e.g.~\cite{nips/HeskesSBC20, aaai/ChenJ20}).

Figure~\ref{fig:linegraph_evaluation_G3} gives more insight into the XAI evaluation practices over time. Years 2014 and 2015 are excluded from this graph, since only 5 and 3 papers respectively were included for those years, leading to unreliable statistics. It confirms that the fraction of papers that quantitatively evaluated their XAI method has slightly increased over the years, whereas the fraction of papers only evaluating with anecdotal evidence shows a decreasing trend. Hence, the evaluation practice in the XAI domain is effectively maturing. 
The number of papers including a user study for evaluation remains however relatively constant over the years at around 20\%. 
Figure~\ref{fig:stacked_barchart_co12} analyzes the number of evaluated Co-12 properties (as introduced in Section~\ref{sec:evaluation_practice}) per paper that introduces an XAI method and quantitatively evaluates it. The leftmost bar shows that 73 papers in our set quantitatively evaluated exactly one Co-12 property, which was often Coherence. The two leftmost bars show that the majority of the papers that introduce an XAI method and quantitatively evaluate it, evaluate one or two Co-12 properties. Coherence and Output-completeness are the Co-12 properties that are evaluated most often, followed by Correctness, Compactness and Covariate complexity. 

\section{Quantitative Evaluation Methods for XAI}
\label{sec:quantitative_evaluation_metrics}

This section presents quantitative evaluation methods that we identified in the 361 included papers. This implies that we do not only consider papers that introduce a method for explaining a machine learning model, but also include the papers that apply or evaluate an XAI method. Our goal is that this section can serve as inspiration and guidance for researchers and practitioners looking for suitable evaluation methods for new or existing XAI methods. 
We focus on functionally-grounded evaluation methods (i.e., \emph{without} user studies) and summarize quantitative evaluation methods \emph{with} user studies in Supplementary Material. 

We clustered all identified quantitative evaluation metrics based on our Co-12 properties and named each of these resulting evaluation methods. 
Table~\ref{tab:evaluation_methods_descriptions_papers} describes each evaluation method we identified, while listing the types of explanations that were mainly related with this method and the papers that applied this evaluation method. Variations and additional information regarding each evaluation method are discussed in Sections~\ref{sec:eval_correctness} to~\ref{sec:eval_controllability}, grouped per Co-12 property. For details regarding the implementation of the specific evaluation metric we refer to the original papers. Table~\ref{tab:evaluation_methods_co12_properties} relates each evaluation method to the corresponding Co-12 properties. The columns indicate what Co-12 properties \emph{can} be measured with the corresponding evaluation method. We note that this does not mean that all cited authors also evaluated all possible properties, but that each related Co-12 property was evaluated in at least one paper. For a thorough and structured evaluation, we argue that it would be good practice to select \emph{multiple} evaluation methods that together cover as much of the Co-12 properties as possible. Such an extensive evaluation would result in a multi-dimensional view on the degree of explainability.

\afterpage{
{
\renewcommand{\arraystretch}{1.4}
{\setstretch{0.85}\selectfont\centering
\begin{longtable}{p{0.81\linewidth}p{0.175\linewidth}}
\caption{Descriptions of automated, quantitative evaluation methods (i.e. without user study), with references to papers that apply this method. The italic text lists the types of explanations that we found mainly related with an evaluation method.}
\label{tab:evaluation_methods_descriptions_papers}\\
\toprule
        \textbf{Name, Description and \textit{Main Explanation Types}} & \textbf{References}\\
\otoprule
\endfirsthead
\caption{Continued}\\
\toprule
        \textbf{Name, Description and \textit{Main Explanation Types}} & \textbf{References}\\
\otoprule
\endhead
\multicolumn{2}{c}{\it \uppercase{Correctness} (Section~\ref{sec:eval_correctness})}\\\midrule
        \textbf{Model Parameter Randomization Check} -- \textit{Feature importance, Heatmap, Localization} \newline Randomly perturb the internals of the predictive model and check that the explanation changes. & \footnotesize{\cite{icml/SixtGL20, ijcai/RosaCN20, kdd/LiangBCBW20, nips/AdebayoMLK20, nips/YehHSIR19}}\\
        \textbf{Explanation Randomization Check} -- \textit{Feature importance, Heatmap} \newline Randomly perturb the explanation (which is built into the predictive model) and check that the output of the predictive model changes. 
         & \footnotesize{\cite{acl/MohankumarNNKSR20, icml/SixtGL20}}\\
        \textbf{White Box Check} -- \textit{Feature importance, Decision Rules, White-box model, Localization} \newline Apply the explanation method to an interpretable white box model and check the correspondence of the explanation with the white box reasoning. & \footnotesize{\cite{icdm/ZhouSC19, icml/LakkarajuAB20, kdd/Jia0RLH19, kdd/Ribeiro0G16, nips/CrabbeZZS20, nips/RamamurthyVZD20, iclr/JinWDXR20}} \\
        \textbf{Controlled Synthetic Data Check} \newline \textit{Feature importance, Heatmap, Prototypes, Localization, White-box model, Graph} \newline Controlled experiment: Create a synthetic dataset such that the predictive model should follow a particular reasoning, known a priori  (important: checking this assumption by e.g. reporting almost-perfect accuracy). Evaluate whether the explanation shows the same reasoning as the data generation process. & \footnotesize{\cite{acl/PruthiGDNL20, acl/SchutzeRP18, iclr/MWT19, icml/ChenSWJ18, icml/KimWGCWVS18, icml/PlumbTST20, ijcai/LiuSH18, nips/AdebayoMLK20, nips/FryeRF20, nips/HoyerMKKF19, nips/IsmailGBF20, nips/LuoCXYZC020, nips/TsangR020, nips/VuT20, nips/YingBYZL19, acl/SydorovaPR19, ijcai/RossHD17,iccv/SubramanyaPP19}} \\
        \textbf{Single Deletion} -- \textit{Feature importance, Heatmap} \newline Delete, mask or perturb a single feature in the input and evaluate the change in output of the predictive model. Measure correlation with explanation's importance score. & \footnotesize{\cite{acl/SerranoS19, cvpr/ZhangYMW19, cvpr/ZhangYMW19, iccv/ChenCHRZ19, iccv/SelvarajuCDVPB17, iccv/SelvarajuLSJGHB19, ijcai/AyushUBLE20, kdd/ChengSHZ19, nips/Alvarez-MelisJ18, nips/FryeRF20, nips/OShaughnessyCCR20}} \\
        \textbf{Incremental Deletion} (or Incremental Addition) -- \textit{Feature importance, Heatmap} \newline One by one delete (or perturb) or add features to the input, based on explanation's order, and measure for each new input the change in output of the predictive model. Report average change in log-odds score, AUC, steepness of curve or number of features needed for a different decision. Compare with random ranking or other baselines. & \footnotesize{\cite{aaai/HuaiWMZ20, aaai/NamGCWL20, acl/ChenZJ20, acl/HanWT20, cvpr/KanehiraTIH19, cvpr/WagnerKGHWB19, cvpr/WangS0H18, cvpr/WuSCZKLT20a, iccv/FongV17, icdm/YangLWH18, iclr/ChenSWJ19, iclr/ChangCGD19, iclr/KindermansSAMEK18, iclr/MWT19, iclr/SinglaPCB20, ijcai/BhattWM20, ijcai/FuscoVVWS19, kdd/PanHLZL20, nips/GhorbaniWZK19, nips/GuoHTXL18, nips/HeoJM19, nips/IsmailGBF20, nips/RaghuGYS17, nips/SchwabK19, nips/YehHSIR19, nips/RamamurthyVZD20, nips/LundbergL17, kdd/LiuHLH18, acl/MohankumarNNKSR20, acl/SerranoS19, nips/HookerEKK19}} \\\midrule
        \multicolumn{2}{c}{\it \uppercase{Output-Completeness} (Section~\ref{sec:eval_completeness})}\\\midrule
        \textbf{Preservation Check} -- \textit{Feature importance, Heatmap, Localization, Text, Prototypes} \newline Giving the explanation (or data based on the explanation) as input to the predictive model should result in the same decision as for the original, full input sample. & \footnotesize{\cite{aaai/YuanCHJ19, acl/BastingsAT19, acl/KumarT20, acl/LiLLLHS20, cvpr/KanehiraH19, icdm/WangCYWW018, iclr/ChangCGD19, icml/0002LA19, icml/ChenSWJ18, ijcai/LuoAPWZYH18, ijcai/RosaCN20, kdd/LiangBCBW20, nips/DhurandharCLTTS18, nips/GuoHTXL18, kdd/YuanTHJ20, nips/YehKALPR20}} \\
        \textbf{Deletion Check} -- \textit{Feature importance, Heatmap, Localization} \newline Giving input \emph{without} explanation's relevant features should result in a different decision by the predictive model than the decision for the original, full input sample. & \footnotesize{\cite{acl/KumarT20, cvpr/PopeKRMH19, iclr/MWT19, ijcai/RosaCN20, kdd/LiangBCBW20, nips/DhurandharCLTTS18}} \\
        \textbf{Fidelity} \newline \textit{Feature importance, Heatmap, Decision Rules, Decision Tree, Prototypes, Text, Localization, White-box model} \newline Measure the agreement between the output of the predictive model and the explanation when applied to the same input sample(s). & \footnotesize{\cite{aaai/WuHPZ0D18, aaai/WuPHKCZ0D20, cvpr/KanehiraH19, cvpr/ZhangYMW19, iccv/ChenCHRZ19, icdm/ZhouSC19, icml/AndersPDMK20, icml/LakkarajuAB20, kdd/Jia0RLH19, kdd/LeW020, kdd/LiuYH18, kdd/PeakeW18, nips/CrabbeZZS20, nips/PedapatiBSD20, nips/PlumbACPXT20, nips/RawalL20, nips/ZhaoP16, www/WuWYLZ18, icdm/ChenL0X19, iclr/YuV17, aaai/DalleigerV20, iclr/TsangCLFZL20, icml/VedantamDLRBP19, aaai/AnnasamyS19}} \\
        \textbf{Predictive Performance} \newline \textit{Feature importance, Heatmap, Decision Rules, Decision Tree, Prototypes, White-box model} \newline Predictive performance of the interpretable model or predictive explanation with respect to the ground-truth data. & \footnotesize{\cite{aaai/HarderBP20, aaai/PolatoA19, aaai/Ribeiro0G18, cvpr/KimGPS20, cvpr/ZhangYMW19, iccv/ChenCHRZ19, icml/AndersPDMK20, nips/CrabbeZZS20, www/WuWYLZ18, aaai/Ribeiro0G18, nips/RawalL20, kdd/GhalwashRO14, iclr/YuV17, ijcai/ZhangR19, kdd/LakkarajuBL16, icdm/YangLWH18, aaai/SilvaFH19, kdd/PeakeW18, icml/PlumbTST20, icdm/LiuMWH0G20}} i.a.\\\midrule
        \multicolumn{2}{c}{\it \uppercase{Consistency} (Section~\ref{sec:eval_consistency})}\\\midrule
        \textbf{Implementation Invariance} -- \textit{Feature Importance} \newline Evaluate whether the explanation method is invariant to specific implementations of the predictive model by validating whether two implementations that give the same output for an input, also get the same explanation. & \footnotesize{\cite{nips/TsengSK20, sigir/FernandoSA19}}\\\midrule
        \multicolumn{2}{c}{\it \uppercase{Continuity} (Section~\ref{sec:eval_continuity})}\\\midrule
        \textbf{Stability for Slight Variations} \newline \textit{Feature importance, Heatmap, Graph, Text, Localization, Decision Rules, White-box model} \newline Measure the similarity between explanations for two slightly different samples. Small variations in the input, for which the model response is nearly identical, should not lead to large changes in the explanation. & \footnotesize{\cite{aaai/GhorbaniAZ19, aaai/GhorbaniAZ19, aaai/TopinV19, acl/CamburuSMLB20, iclr/PuriVGKDK020, iclr/SinghMY19, icml/Boopathy0ZLCCD20, kdd/LiangBCBW20, nips/Alvarez-MelisJ18,  icml/LakkarajuAB20, kdd/ChuHHWP18, icml/PlumbTST20, icml/SundararajanN20, nips/DombrowskiAAAMK19, nips/HeoJM19, nips/PlumbACPXT20, nips/YehHSIR19, sigir/VermaG19, nips/PlumbMT18, acl/LiLLLHS20}}\\
        \textbf{Fidelity for Slight Variations} -- \textit{Decision Rules, White-box model} \newline Measure the agreement between interpretable predictions for original and slightly different samples: an explanation for original input $x$ should accurately predict the model's output for a slightly different sample $x'$. & \footnotesize{\cite{icml/LakkarajuAB20, nips/PlumbMT18}}\\
        \textbf{Connectedness} -- \textit{Prototypes, Representation Synthesis} \newline Measure how connected a counterfactual explanation is to samples in the training data: ideally, the counterfactual is not an outlier, and there is a continuous path between a generated counterfactual and a training sample. & \footnotesize{\cite{ijcai/LaugelLMRD19,www/PawelczykBK20, ijcai/KanamoriTKA20}}\\\midrule
        \multicolumn{2}{c}{\it \uppercase{Contrastivity} (Section~\ref{sec:eval_contrastivity})}\\\midrule
        \textbf{Target Sensitivity} -- \textit{Heatmap} \newline The explanation for a particular target or model output (e.g. class) should be different from an explanation for another target. & \footnotesize{\cite{cvpr/PopeKRMH19, iccv/SubramanyaPP19, icml/NieZP18, icml/SixtGL20, cvpr/WagnerKGHWB19, icml/WangZB19}}\\
        \textbf{Target Discriminativeness} -- \textit{Disentanglement, Representation Synthesis, Text} \newline The explanation should be target-discriminative such that \emph{another model} can predict the right target (e.g.~class label) from the explanation, in either a supervised or unsupervised fashion. & \footnotesize{\cite{cvpr/WuSCZKLT20a, icdm/JhaWXZ18, nips/CamburuRLB18, cvpr/WangS0H18, iclr/FortuinHLSR19, iclr/KindermansSAMEK18, icml/VedantamDLRBP19, icml/VoynovB20, iclr/SinglaPCB20}} \\
        \textbf{Data Randomization Check} -- \textit{Feature importance, Heatmap, Localization} \newline Randomly change labels in a copy of the training dataset, train a model on this randomized dataset and check that the explanations for this model on a test set are different from the explanations for the model trained on the original training data. & \footnotesize{\cite{ijcai/RosaCN20, kdd/LiangBCBW20, nips/AdebayoMLK20}} \\
        \midrule
        \multicolumn{2}{c}{\it \uppercase{Covariate Complexity} (Section~\ref{sec:eval_complexity})}\\\midrule
        \textbf{Covariate Homogeneity} \newline \textit{Prototypes, Disentanglement, Localization, Heatmap, Representation Synthesis} \newline Evaluate how consistently a covariate (i.e. feature) in an explanation represents a predefined human-interpretable concept. &\footnotesize{\cite{aaai/ZhangCSWZ18, iccv/SunRS19, nips/LakkarajuL16, nips/MuA20, cvpr/ZhangWZ18a, iclr/FortuinHLSR19, acl/EskenaziLZ18, cvpr/EsserRO20, cvpr/FongV18, cvpr/LiuLZKWBRC20, cvpr/ShenGTZ20, iccv/YinTLS019, icml/AdelGW18, icml/ShiZM020, ijcai/BeyazitTYT020, kdd/GuoZQWSY20, cvpr/BauZKO017, nips/TsangLPML18, iclr/ZhengPBH19, nips/HsuZG17, acl/FysheTMM14, cvpr/KanehiraTIH19, iclr/SinglaPCB20, iccv/WorrallGTB17}}\\
        \textbf{Covariate Regularity} -- \textit{Decision Rules, Feature Importance} \newline Evaluate the regularity of an explanation by measuring its Shannon entropy, in order to quantify how noisy the explanation is and how easy it is to memorize the explanation.  & \footnotesize{\cite{iclr/YuV17, nips/TsengSK20}}\\\midrule
        \multicolumn{2}{c}{\it \uppercase{Compactness} (Section~\ref{sec:eval_compactness})}\\\midrule
        \textbf{Size} \newline \textit{Feature importance, Heatmap, Decision Rules, Decision Tree, Prototypes, Text, Graph, Localization, White-box model, Representation Synthesis} \newline Total size (absolute) or sparsity (relative) of the explanation. & \footnotesize{\cite{aaai/IgnatievNM19, aaai/PolatoA19, aaai/Ribeiro0G18, aaai/RustamovK18, aaai/ShakerinG19, aaai/TopinV19, aaai/WuHPZ0D18, aaai/WuPHKCZ0D20, aaai/ZhangCWZ17, acl/SchutzeRP18, cvpr/PopeKRMH19, cvpr/WangS0H18, iccv/SunRS19, icdm/WangCYWW018, icdm/YooS19, icml/ChalasaniC00J20, icml/PlumbTST20, icml/Wang19, ijcai/AlbiniRBT20, kdd/Ribeiro0G16, nips/CrabbeZZS20, nips/KimKK16, nips/KimL19, nips/LageRGKD18, nips/PlumbACPXT20, nips/PlumbMT18, nips/RawalL20, nips/Wang18, sigir/TaoJWW19, sigir/VermaG19, icdm/ChenL0X19, kdd/GhalwashRO14, ijcai/ZhangR19, icdm/KarlssonRPG18, aaai/DalleigerV20, kdd/LakkarajuBL16,www/WuWYLZ18, acl/MohankumarNNKSR20, acl/LiLLLHS20}} \\
        \textbf{Redundancy} -- \textit{Feature importance, Decision Rules, Text, White-box model} \newline Calculate the redundancy or overlap between parts of the explanation. & \footnotesize{\cite{nips/TsangR020, kdd/LakkarajuBL16, kdd/LeW020}} \\
        \textbf{Counterfactual Compactness} -- \textit{Prototypes, Representation Synthesis, Text} \newline Given a counterfactual explanation showing what needs to be changed in the input in order to change the prediction of the predictive model, measure how \emph{much} needs to be changed. & \footnotesize{\cite{icdm/KarlssonRPG18, icml/GoyalWEBPL19, ijcai/KanamoriTKA20, kdd/LeW020, www/PawelczykBK20, nips/ZhangSS18, ijcai/AlbiniRBT20, kdd/TolomeiSHL17}}\\\midrule
        \multicolumn{2}{c}{\it \uppercase{Composition} (Section~\ref{sec:eval_composition})}\\\midrule
        \textbf{Perceptual Realism} -- \textit{Representation Synthesis, Text} \newline Measure how realistic a generated explanation is compared to real, original samples. & \footnotesize{\cite{cvpr/EsserRO20, iclr/SinglaPCB20, nips/CamburuRLB18}}\\
        \midrule
        \multicolumn{2}{c}{\it \uppercase{Confidence} (Section~\ref{sec:eval_confidence})}\\\midrule
        \textbf{Confidence Accuracy} -- \textit{Feature Importance, Prototypes} \newline Measure the accuracy of confidence/uncertainty estimates if these are present in the explanation.  & \footnotesize{\cite{nips/SchwabK19, kdd/GhalwashRO14}}\\
        \midrule
        \multicolumn{2}{c}{\it \uppercase{Context} (Section~\ref{sec:eval_context})}\\\midrule
        \textbf{Pragmatism} -- \textit{Decision Rules, Representation Synthesis} \newline The cost or degree of difficulty for a user to act upon suggestions by a counterfactual explanation that explains what the user should change to attain a particular outcome by the predictive model. & \footnotesize{\cite{www/PawelczykBK20, nips/RawalL20}}\\
        \textbf{Simulated User Study} -- \textit{Feature Importance, Localization} \newline Create a synthetic dataset such that the utility of explanations for user-relevant tasks can be automatically evaluated. & \footnotesize{\cite{kdd/Ribeiro0G16, iclr/SinglaPCB20}} \\
        \midrule
        \multicolumn{2}{c}{\it \uppercase{Coherence} (Section~\ref{sec:eval_coherence})}\\\midrule
        \textbf{Alignment with Domain Knowledge} \newline \textit{Feature importance, Heatmap, Localization, Text, Disentanglement, Prototypes} \newline Compare the generated explanation with a `ground-truth' expected explanation based on domain knowledge. & \scriptsize{\cite{aaai/NamGCWL20, aaai/WickramanayakeH19, acl/MohankumarNNKSR20, acl/WuCKL20, cvpr/ChengRCZ20, cvpr/HuangL20, cvpr/JakabGBV20, cvpr/MascharkaTSM18, cvpr/WangV20, cvpr/ZengLSSYCU19, iccv/FongV17, iccv/SelvarajuCDVPB17, iclr/ChangCGD19, icml/GoyalWEBPL19, icml/JinBJ20a, ijcai/PanLLZ20, kdd/DuLSH18, nips/BassSSTSR20, nips/HoyerMKKF19, nips/WangN19, cvpr/BauZKO017, cvpr/ZhangWZ18a, kdd/LiuHLH18, aaai/PatroAN20, aaai/TuHW0HZ20, acl/AtanasovaSLA20, acl/BastingsAT19, acl/JiangJCB19, acl/LiuYW19, acl/RajaniMXS19, acl/SchutzeRP18, acl/ShahbaziFGT20, acl/SubramanianBGWS20, cvpr/ChuangL0F18, cvpr/LiuLZKWBRC20, cvpr/ParkHARSDR18, cvpr/WagnerKGHWB19, cvpr/XuYGLWLV20, iccv/PatroLPN19, iccv/SelvarajuLSJGHB19, icdm/JhaWXZ18, iclr/JinWDXR20, iclr/PuriVGKDK020, icml/AndersPDMK20, ijcai/ChenW0PSAC20, ijcai/ChenW0WBWC19, ijcai/LeL20, nips/CamburuRLB18, nips/HeoLKLKYH18, nips/TsangR020, nips/TsengSK20, sigir/ChenCXZ0QZ19, sigir/ChenYYH0020, sigir/VermaG19, www/ChenZLM18, www/SunWZFHW20, cvpr/DongSZZ17, iclr/SinglaPCB20, cvpr/KanehiraH19, icdm/WangCYWW018, icml/EtmannLMS19, ijcai/RosaCN20}}\\
        \textbf{XAI Methods Agreement} -- \textit{Feature importance, Heatmap, Localization} \newline Quantitatively compare explanations from different XAI methods and evaluate their agreement. & \footnotesize{\cite{aaai/NamGCWL20, aaai/WangLRRXS20, aaai/WickramanayakeH19, acl/HanWT20, acl/MohankumarNNKSR20, iclr/JinWDXR20, icml/AnconaOG19, ijcai/FuscoVVWS19, kdd/ShuCW0L19, nips/0001GCIN20}}\\
        \midrule
        \multicolumn{2}{c}{\it \uppercase{Controllability} (Section~\ref{sec:eval_controllability})}\\\midrule
        \textbf{Human Feedback Impact} -- \textit{Interactive Explanation, Text} \newline Measure the improvement of explanation quality after human feedback, where the user is seen as a system component. &\footnotesize{ \cite{ijcai/ChenW0PSAC20, cvpr/DongSZZ17}}\\
        \bottomrule
\end{longtable}
}
}}
\normalsize

\afterpage{
\renewcommand{\arraystretch}{1.0}
\begin{table}
    \centering
    \caption{Identified quantitative methods to evaluate explanations without user studies, related to their Co-12 properties. Bold check mark indicates prominent Co-12 property. Superscript R and O indicate Reasoning-completeness resp. Output-completeness.}
    \label{tab:evaluation_methods_co12_properties}
    \begin{tabular}{l|cc*{5}{cc}}
        \textbf{Name} &\rot{Correctness} &\rot{Completeness} &\rot{Consistency} & \rot{Continuity} &\rot{Contrastivity} & \rot{Covariate Compl.} & \rot{Compactness} &\rot{Composition} & \rot{Confidence} & \rot{Context} & \rot{Coherence} & \rot{Controllability}\\
        \toprule
        Model Parameter Randomization Check &\boldcheckmark&&&&&&&&&&&\\
        \rowcolor{verylightgray} Explanation Randomization Check &\boldcheckmark&&&&&&&&&&&\\
        White Box Check & \boldcheckmark &\checkmark\textsuperscript{\scriptsize{R}}& & & & & &&&&&\\
        \rowcolor{verylightgray}Controlled Synthetic Data Check & \boldcheckmark & \checkmark\textsuperscript{\scriptsize{R}} &&&&&&&&&\checkmark&\\
        Single Deletion & \boldcheckmark &  & & & & & &&&&&\\
        \rowcolor{verylightgray}Incremental Deletion (or Incremental Addition) & \boldcheckmark & \checkmark\textsuperscript{\scriptsize{O}} & & & &  & \checkmark&&&&&\\
        Preservation Check && \boldcheckmark\textsuperscript{\scriptsize{O}}& & & & & &&&&&\\
        \rowcolor{verylightgray}Deletion Check && \boldcheckmark\textsuperscript{\scriptsize{O}}& & & & & &&&&&\\
        Fidelity && \boldcheckmark\textsuperscript{\scriptsize{O}}& & & & & &&&&&\\
        \rowcolor{verylightgray}Predictive Performance && \boldcheckmark\textsuperscript{\scriptsize{O}}& & & & & &&&&&\\
        Implementation Invariance &&&\boldcheckmark& & & & &&&&&\\
        \rowcolor{verylightgray}Stability for Slight Variations &&&&\boldcheckmark&&&&&&&&\\
        Fidelity for Slight Variations &&\checkmark\textsuperscript{\scriptsize{O}}&&\boldcheckmark&&&&&&&&\\
        \rowcolor{verylightgray}Connectedness &&&&\boldcheckmark&&&&&&&&\\
        Target Sensitivity &&&&&\boldcheckmark&&&&&&&\\
        \rowcolor{verylightgray}Target Discriminativeness &&&&&\boldcheckmark&&&&&&&\\
        Data Randomization Check &&&&&\boldcheckmark&&&&&&&\\
        \rowcolor{verylightgray}Covariate Homogeneity &&&&&&\boldcheckmark&&&&&\checkmark&\\
        Covariate Regularity &&&&&&\boldcheckmark&&&&&&\\
        \rowcolor{verylightgray}Size &  & & & & & & \boldcheckmark &&&&&\\
        Redundancy &  & & & & & & \boldcheckmark &&&&&\\
        \rowcolor{verylightgray}Counterfactual Compactness &&&&&\checkmark&&\boldcheckmark&&&&&\\
        Perceptual Realism &  & & & & & & &\boldcheckmark &&&&\\
        \rowcolor{verylightgray}Confidence Accuracy & \checkmark & \checkmark\textsuperscript{\scriptsize{O}} & & & & & & &\boldcheckmark&&&\\
        Pragmatism &  & & & &\checkmark & & \checkmark &&&\boldcheckmark&&\\
        \rowcolor{verylightgray}Simulated User Study &  & & & & & & &&&\boldcheckmark&&\\
        Alignment with Domain Knowledge &  & & & & & & &\checkmark&&&\boldcheckmark&\\
        \rowcolor{verylightgray}XAI Methods' Agreement &  & & & & & & &&&&\boldcheckmark&\\
        Human Feedback Impact & & &&&&&&&&&\checkmark&\boldcheckmark\\
        \bottomrule
    \end{tabular}
    \label{tab:evaluation_methods_bigtable}
\end{table}}
\normalsize

\subsection{Functionally Evaluating Correctness}
\label{sec:eval_correctness}
The correctness property addresses to what extent the explanation is faithful to the predictive model it explains. Important to emphasize here is that an explanation that looks reasonable to a user is not guaranteed to also be truthfully reflecting the reasoning of the model~\cite{Jacovi_towards_2020}. Checking the correctness of an explanation with respect to the predictive model $f$ is therefore different from the plausibility to the user. 

The \textbf{Model Parameter Randomization Check} was introduced by Adebayo et al.~\cite{adebayo_sanity_2018} as ``sanity check'' for the faithfulness and sensitivity of the explanation to predictive model $f$. Perturbation of model $f$ can be done by randomizing parameters or re-initializing weights, after which the explanation is expected to change. If the explanation after randomization is the same as the original explanation, then the explanation is not sensitive to $f$ and hence not correct w.r.t. reasoning of model $f$. (Our recommendation is to do multiple randomization runs to ensure that the two explanations are not accidentally similar.) However, if the explanations are different, it is not a guarantee that the original explanation is fully correct. It is therefore presented as a sanity check:  the sensitivity of $e$ to parametrization changes in predictive model $f$ is a necessary but not sufficient condition for correctness.
A related method is the \textbf{Explanation Randomization Check} which is applicable to explanations that are built into predictive model $f$, such as attention and backpropagated relevance vectors. Permuting and randomizing the explanation within the model should change the model's output, and therefore evaluates the sensitivity of the explanation to the predictive model. 

The \textbf{White Box Check} is designed to evaluate correctness by training a white-box model as predictive model and applying the explanation method to the white-box model as if it was a black box. Since the reasoning of a white-box model is known, the explanation can subsequently be compared with the true reasoning in order to evaluate how closely the explanation resembles the model's reasoning. Therefore, also Reasoning-completeness is evaluated since the `golden' reasoning is known and can be compared to the degree of information in the explanation. Instead of a fully transparent model, Ramamurthy et al.~\cite{nips/RamamurthyVZD20} use a random forest and compare the explanation with feature importance scores output by the forest based on established methodology for these type of models. 

The \textbf{Controlled Synthetic Data Check} is useful for evaluating explanations for black box models. By designing a dataset in such a way that with relatively high confidence one could say that predictive model $f$ reasons in a particular way, `gold' explanations can be created that follow the data generation process. Subsequently, the agreement of the generated explanations with these true explanations can be measured. For example, Oramas et al.~\cite{iclr/MWT19} generate an artificial image dataset of flowers where the color is the discriminative feature between classes. They subsequently compare their explanations with the (location of the) discriminative, colored area. Recently, a set of synthetic benchmarks for XAI was published
~\cite{liu2021synthetic}. The controlled synthetic data check also implies that Reasoning-completeness is evaluated, but then only on discriminatory feature level and not necessarily how the features are aggregated by the model. 
An important prerequisite for the Controlled Synthetic Data Check is that it should be reasonable to assume that the black box has learned the intended reasoning. Since the predictive model is a black box, reporting the task accuracy of model $f$ or using other checks on the output is good practice to validate that it is safe to assume that the model has picked up the intended reasoning. 

The correctness of real-valued feature importance scores and heatmaps is often evaluated by removing, perturbing or masking features from the input and measuring how that affects the (confidence of the) output of predictive model $f$. The \textbf{Single Deletion} method evaluates the change in output when removing or perturbing one feature and compares that with the explanation's importance score. In a correct explanation, the explanation's feature importance score should be proportional to the shift in output distribution. In other words, the feature with the highest importance score should also lead to the biggest change in output from $f$, and features with a low importance should not result in a significant change. The Single Deletion method also allows to check for specific properties, such as the ``null attribute'' indicating that omitting a feature that has no effect on the output of the model, should have an importance score of zero~\cite{ijcai/LabreucheF18}. 
Features can also be removed one by one in an iterative, incremental fashion: \textbf{Incremental Deletion}. Given the exponential number of possible subsets, features are often removed incrementally in either descending order (i.e. most important feature first, then top-2 most important features, etc.) or ascending order (least important or most unimportant first). Due to the high computational costs of iterative removal, some authors (e.g.~\cite{acl/HanWT20, ijcai/FuscoVVWS19, nips/SchwabK19}) do not evaluate output of $f$ at each iteration, but only evaluate it for specific subsets, such as removing the top-k most influential and least influential features. Authors using Incremental Deletion often refer to work of Shrikumar et al.~\cite{pmlr-v70-shrikumar17a} who calculate the difference in log-odds scores by $f$, and Samek et al.~\cite{samek_evaluating_2017} who measure the Area over the Perturbation Curve when perturbing a region in an image. Analyzing the area or steepness of a curve however assumes that the majority of the importance is placed on only a few features. Although this is typically the case for softmax scores~\cite{acl/SerranoS19}, there might be cases where this assumption is invalid. It is therefore good practice to compare the curve with other baselines, such as a random ranking. 
Instead of starting with the full input and incrementally removing features, some authors start with an `empty' input and incrementally \emph{add} features: \textbf{Incremental Addition} (e.g.~\cite{cvpr/WuSCZKLT20a}).

\begin{figure}[b!]
    \centering
    \includegraphics[width=0.82\linewidth]{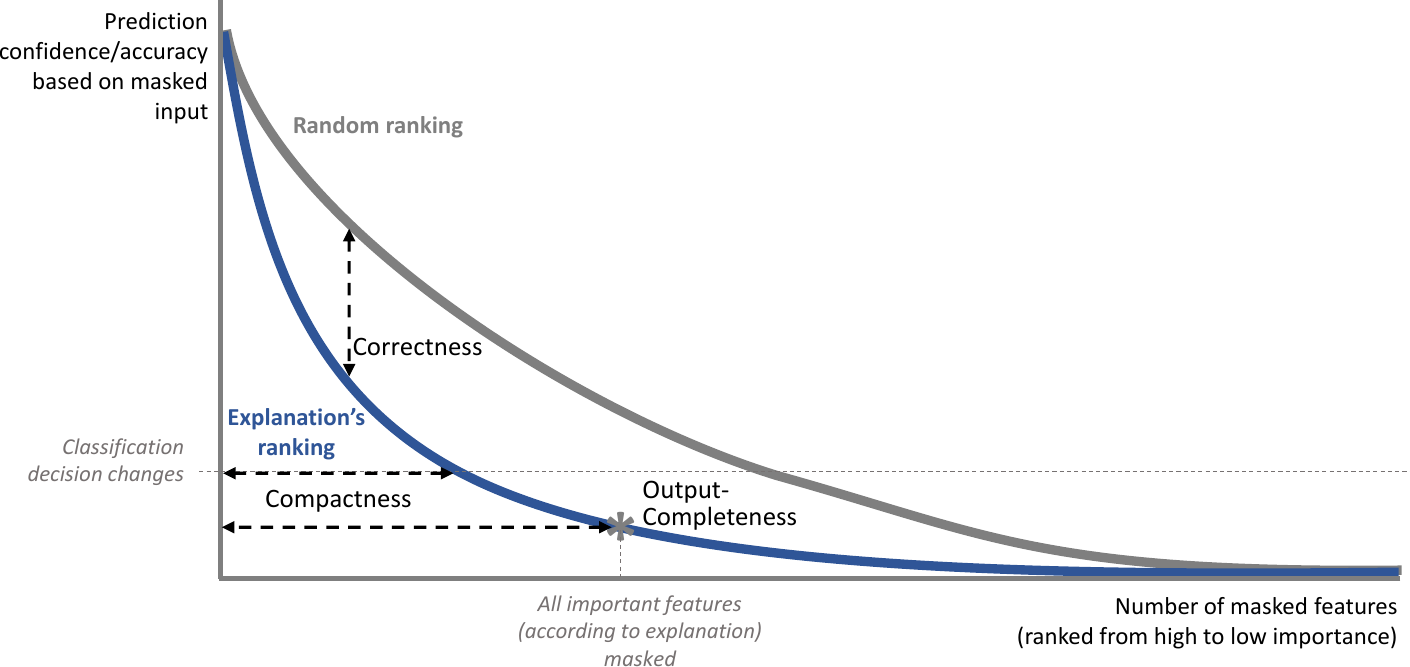}
    \caption{The evaluation curve for the Incremental Deletion method to evaluate correctness, output-completeness and compactness. An explanation is output-complete if the model's decision changes when all important features are removed. Compactness can be measured at point of output-completeness, or at model's decision boundary. Explanation is correct when deleting the most important features leads to the biggest drop in prediction output compared to randomly deleting features.}
    \label{fig:curve_correctness_completeness_compactness}
\end{figure}

\textbf{Incremental Deletion/Addition} can include the evaluation of output-completeness. Where \emph{correctness} evaluates whether the importance score value is correct, \emph{output-completeness} evaluates whether the set of important features is sufficient to explain the output of model $f$. For example: an explanation showing one relevant pixel might be correct but is probably not output-complete, and an explanation showing the full input image is output-complete but the importance score of pixels is probably incorrect.
Output-completeness is evaluated for Incremental Deletion at the point where all important features are removed. An output-complete explanation should result in a wrong decision by model $f$ when all features in the explanation are removed from the input. For Incremental Addition, the output of the model when only the important features are present should be similar to the output for the full input (see also Preservation Check and Deletion Check). 
Besides using Incremental Deletion/Addition to evaluate correctness and output-completeness, compactness can be evaluated by counting how many features of the explanation need to be removed, perturbed or added in order to change the decision of the predictive model~\cite{acl/MohankumarNNKSR20, acl/SerranoS19, iccv/FongV17, icdm/KarlssonRPG18, ijcai/AlbiniRBT20, iclr/ChangCGD19}. The motivation is that an explanation is easier to contemplate at once when this minimal set is small. Figure~\ref{fig:curve_correctness_completeness_compactness} visualizes the evaluation curve for Incremental Deletion and the points on this curve where output-completeness and compactness can be evaluated.

The criticism on \textbf{Single Deletion} and \textbf{Incremental Deletion} is that deletion of features, such as setting them to zero, can lead to out-of-distribution samples~\cite{iclr/ChangCGD19, nips/HookerEKK19, nips/IsmailGBF20}. A solution would be to train a new model on the modified data. For example, Frye et al.~\cite{nips/FryeRF20} train separate classifiers for different feature subsets, and Hooker et al.~\cite{nips/HookerEKK19} remove the $k\%$ most important features and then retrain the predictive model. This however implies that the correctness with respect to the \emph{original} model is not evaluated. Chang et al.~\cite{iclr/ChangCGD19} solve the out-of-distribution issue by applying a Generative Adversarial Network (GAN) to fill in deleted features, and Ismail et al.~\cite{nips/IsmailGBF20} replace masked features with values from the original data distribution. Recently, it was found that also the \emph{shape} of a mask could leak class information to the model~\cite{pmlr-v162-rong22a} and various approaches are presented to circumvent these OoD and shape-leakage issues, e.g.~\cite{pmlr-v162-rong22a,hase_nips_21_ood_xai}. 

\subsection{Functionally Evaluating Output-Completeness}
\label{sec:eval_completeness}
The \textbf{Preservation Check} and \textbf{Deletion Check}, following terminology of~\cite{cvpr/WagnerKGHWB19}, evaluate the output-completeness of the explanation: i.e., does the explanation hold enough information to explain the output of model $f$. Explanations evaluated in this way are usually outcome explanations or model inspections. The methodology is similar to the Incremental Deletion and Incremental Addition method, but instead of incrementally deleting features, the whole explanation is removed from the input at once (analogously for preservation). This visualizes a single point on the incremental deletion curve, as shown in Figure~\ref{fig:curve_correctness_completeness_compactness}. These checks are measured by calculating the change in accuracy or confidence of the predictive model (hence, \emph{output}-completeness). Ideally, the accuracy of $f$ for the Deletion Check should lead to a significant drop in accuracy whereas the accuracy after the Preservation Check should stay similar. Nam et al.~\cite{aaai/NamGCWL20} note that a small change in accuracy can be inevitable due to distortion of the input (such as color and shape in images) which can result in unpredictable noise that affect the model's output. It is therefore good practice to only consider significant changes by comparing with random deletions/preservations.

\textbf{Fidelity} measures the agreement between the output of predictive model $f$ and the explanation when applied to the input, and therefore evaluates how well the explanations mimic the output of model $f$~\cite{guidotti2018survey,trepan_craven_1995}.  
It can be applied to model explanations and outcome explanations. For example, when a decision tree is trained to generate the same predictions as a neural network and therefore acts as an explainable surrogate model, the predictions of the decision tree can be compared with the predictions of the neural network~\cite{trepan_craven_1995}. 
Fidelity is often defined as the fraction of data samples for which predictive model $f$ and an explanation make the same decision, but can also be reported as approximation error by calculating the average absolute difference or mean squared error~(e.g.~\cite{icdm/ZhouSC19, icml/AndersPDMK20}). Others use slight variations on fidelity such as the Kullback–Leibler divergence between the outputs~(e.g.~\cite{icml/AndersPDMK20, icdm/ChenL0X19, iclr/YuV17}), conditional entropy~\cite{iccv/ChenCHRZ19}, correlation~\cite{ijcai/ZhangR19}, likelihood comparison~\cite{aaai/DalleigerV20} or evaluating how adding explanations to the model would improve prediction performance~\cite{iclr/TsangCLFZL20}.
Le et al.~\cite{kdd/LeW020} also introduced a metric called ``influence'' which combines fidelity with information gain and compactness. We emphasize that fidelity is \emph{not} evaluating correctness, although sometimes presented as such. Since only the outputs of the model and the explanation are compared, there is no guarantee that the explanation follows the same reasoning as $f$. For example, Anders et al.~\cite{icml/AndersPDMK20} theoretically showed that for any classifier, one can always construct another classifier that gives the same output as the original classifier for all data instances, but has arbitrarily manipulated explanation maps. 

When predictive model $f$ equals explanation method $e$, the explanation is a transparent box design (such as decision rules, decision tree or another white-box model that is explainable \emph{and} makes the predictions). Hence, fidelity is not applicable for transparent boxes. Instead, the \textbf{Predictive Performance} of the transparent box model with respect to the ground-truth task data can be evaluated, such as classification accuracy. In case of $f\neq e$, the predictive performance of $f$ is not directly related to the explanation quality and therefore not included as explanation evaluation metric (although it is generally good practice to report task accuracy and related metrics for evaluating $f$, since this can influence the perceived coherence of the explanation). However, some authors~(e.g.~\cite{aaai/HarderBP20,iccv/ChenCHRZ19}) evaluate output-completeness by comparing the accuracy of their explanation method with the accuracy of predictive model $f$. The decrease in accuracy then quantifies output-completeness. This approach however does not capture whether $e$ and $f$ misclassify the same test samples. Rather than comparing the predictions of $e$ and $f$ with the ground-truth labels, it is therefore more informative to compare them with each other (fidelity). 
Another implementation of Predictive Performance is coverage, which quantifies the fraction of samples to which the explanation applies. The higher the coverage, the more output-complete the explanation is and the higher the predictive performance can be. In case of outcome explanations, a set of explanations can be generated for the training set, after which the coverage for this set is evaluated on the test set. With decision rules for example, coverage would measure the fraction of instances that is classified by at least one rule in the rule set~\cite{aaai/Ribeiro0G18, ijcai/ZhangR19, kdd/LakkarajuBL16}. 

\subsection{Functionally Evaluating Consistency}
\label{sec:eval_consistency}
Consistency evaluates whether identical inputs have identical explanations. In practice, this can  address to what extent the explanation method is deterministic. Determinism of an explanation method is usually a design choice and therefore only (implicitly) qualitatively discussed. 
A quantitative method related to consistency is \textbf{Implementation Invariance} which states that two models that give the same outputs for all inputs (regardless of their internal implementation) should have the same explanations. Although definitions differ slightly between authors, Tseng et al.~\cite{nips/TsengSK20} evaluated Implementation Invariance by computing the Jaccard similarity between feature importance scores across random initializations of the predictive model. Fernando et al.~\cite{sigir/FernandoSA19} use a special version of Implementation Invariance by specifically focusing on `reference inputs' for the DeepSHAP explanation method. They explain that a plain black image is a standard reference input to compute relative importance for image classification, and analyze the sensitivity of explanations to different reference inputs for retrieval tasks. 

\subsection{Functionally Evaluating Continuity}
\label{sec:eval_continuity}
Continuity addresses the generalizability of explanations and can be measured with the 
\textbf{Stability for Slight Variations} method, which measures the similarity between explanations for an original input sample and a slightly different version of this sample. Alvarez-Melis and Jaakkola~\cite{nips/Alvarez-MelisJ18}, among others, introduced the term \emph{stability}, but also \emph{sensitivity}~\cite{nips/YehHSIR19, kdd/LiangBCBW20} and \emph{robustness}~\cite{iclr/PuriVGKDK020, iclr/SinghMY19} are used. 
Most authors add a small amount of noise to an original input sample or otherwise slightly perturb a sample. A few others evaluate stability between two original samples that are similar, e.g. by using a local neighborhood criterion~\cite{nips/PlumbACPXT20}. Similarity between explanations can be quantified with various metrics, often dependent on the type of explanation. Examples include rank order correlation~\cite{aaai/GhorbaniAZ19, iclr/SinghMY19, icml/Boopathy0ZLCCD20, sigir/VermaG19}, top-k intersection~\cite{aaai/GhorbaniAZ19}, cosine similarity~\cite{kdd/ChuHHWP18}, rule match~\cite{icml/LakkarajuAB20}, normalized distance~\cite{nips/Alvarez-MelisJ18} and the structural similarity index (SSIM)~\cite{nips/DombrowskiAAAMK19}. Nie et al.~\cite{icml/NieZP18} show however that one should always check whether the output of the predictive model stays the same for a slightly perturbed input, before evaluating the similarity of explanations. 

Instead of comparing the explanations, others evaluate the \textbf{Fidelity for Slight Variations} by comparing the \emph{predictions} of $f$ for original and slightly perturbed inputs. The reasoning is that an explanation, which should be a predictive model such that $e=f$, for original input $x$ should accurately predict the model's output for a slightly different sample $x'$~\cite{nips/PlumbMT18}. Lakkaraju et al.~\cite{icml/LakkarajuAB20} argue that explanations should have high fidelity on both the original input data and on slightly shifted input to ensure robustness of the explanations. 

Continuity can also be evaluated for counterfactual explanations, which address hypothesized events that did not occur in reality~\cite{stepin2021survey}, to answer questions as ``How would the prediction have been if the input had been different?''~\cite{molnar2020interpretable}. The explanation can be a (generated) data instance that is predicted to belong to a different class~\cite{ijcai/LaugelLMRD19}, such as `\textit{your loan would have been accepted if your income would be 10k higher}'. 
Laugel et al.~\cite{ijcai/LaugelLMRD19} and Pawelczyk et al.~\cite{www/PawelczykBK20} evaluate their counterfactual explanations by calculating \textbf{Connectedness}. They argue that generated counterfactual samples should be ``justified''~\cite{ijcai/LaugelLMRD19}, meaning that there is a continuous path from the counterfactual to a sample in the ground-truth training data. Closely related is requiring that the counterfactual explanation is in proximity of actual instances to prevent that the explanation is an outlier~\cite{www/PawelczykBK20, ijcai/KanamoriTKA20}. Connectedness addresses the lack of robustness and unknown generalization ability of some predictive models. Requiring connectedness would prevent generating counterfactual explanations that are based on artifacts which are correct with respect to a model that is not robust, but not near an instance from the training data. The latter can be undesirable for the Co-12 properties Context and Coherence. 

\subsection{Functionally Evaluating Contrastivity}
\label{sec:eval_contrastivity}
Contrastivity addresses the discriminativeness of explanations with respect to a ground-truth label or other target. Therefore the \textbf{Target Sensitivity} evaluation method captures ``the intuition that class-specific features highlighted by an explanation should differ between classes''~\cite{cvpr/PopeKRMH19}. Interestingly, we found that Target Sensitivity was only evaluated for heatmaps. Sixt et al.~\cite{icml/SixtGL20} confirmed the relevance of this evaluation method by proving that some attribution methods can converge to class-insensitive explanations. Moreover, Adebayo et al.~\cite{adebayo_sanity_2018} showed that visual inspection can favor plausible heatmaps which are not target-sensitive to the underlying reasoning of the model. They illustrated this by revealing the high similarity between edge detectors for image data and explanatory saliency maps.
Ideally, explanations are target-sensitive and should differ between targets and, in case of image data, should not be static edge detectors~\cite{icml/NieZP18,adebayo_sanity_2018}. This can be checked by comparing explanations for different targets or output logits~\cite{cvpr/PopeKRMH19, icml/NieZP18, icml/SixtGL20, nips/ZhangSS18} or explanations before and after an adversarial attack~\cite{icml/NieZP18, iccv/SubramanyaPP19}. An adversarial attack fools the predictive model $f$ such that it makes a different prediction for a slightly perturbed input. A different prediction should then also lead to a different explanation.  Target Sensitivity can be measured with L1, L2 or Hamming distance between the explanations~\cite{cvpr/PopeKRMH19, icml/NieZP18, nips/ZhangSS18}, histogram intersection~\cite{iccv/SubramanyaPP19} or the structural similarity index measure (SSIM) between two heatmaps~\cite{icml/SixtGL20}. In all these cases, a large difference between the explanations is desired. Wagner et al.~\cite{cvpr/WagnerKGHWB19} go a step further and argue that explanations should be empty for targets for which there is no evidence in the input sample (e.g.~a class that is visually not present in an image). They generate an explanation for the least likely class and compute the fraction of explanations that was not empty~\cite{cvpr/WagnerKGHWB19}. 

Another desirable property regarding contrastivity is high \textbf{Target Discriminativeness}, since that implies a good ``informativeness for a downstream prediction task''~\cite{iclr/FortuinHLSR19}. To evaluate the target-discriminativeness, either an external classifier is trained on predicting the right target given the explanation~\cite{cvpr/WuSCZKLT20a, icdm/JhaWXZ18, nips/CamburuRLB18}, or a cluster method is applied on the explanations~\cite{cvpr/WangS0H18, iclr/FortuinHLSR19}. Explanations in these papers are often interpretable representations or text. The performance of the classifier or clustering method is evaluated against ground-truth targets. 
Kindermans et al.~\cite{iclr/KindermansSAMEK18} evaluate \textbf{Target Discriminativeness} slightly different by evaluating how much of the target can be reconstructed when the explanation is \emph{removed} from the input. 

The \textbf{Data Randomization Check}, introduced by Adebayo et al.~\cite{adebayo_sanity_2018}, is a sanity check for the ``sensitivity of an explanation method to the relationship between instances and labels'', and has the advantage that it is model-agnostic. Ideally, explanations explain the mapping between input and output that the predictive model has learned. If a model is successfully trained on a dataset with random labels, which has been shown to be possible with deep neural networks~\cite{zhang_understanding_randomizedlabels_2021}, it has learned the data generation process by memorizing the random labels. The accuracy on an unseen test set will never be better than random guessing~\cite{adebayo_sanity_2018}. Therefore, for a given test sample, the explanation of a model trained on randomized data should be different from an explanation of a model trained on the original dataset. Randomizing the underlying data generation process changes the target function the model is trying to learn, which should imply a change in the explanation.

\subsection{Functionally Evaluating Covariate Complexity}
\label{sec:eval_complexity}
Covariate complexity is concerned with the use of human-understandable concepts to explain features (i.e. covariates) and their interactions.
It can be qualitatively addressed by motivating a design choice, such as using a bag of words instead of uninterpretable word embeddings as covariates~\cite{kdd/Ribeiro0G16}, applying element-wise feature mappings which are claimed to be more interpretable~\cite{aaai/LanG19} or selecting a predictive model that satisfies a monotonicity requirement~\cite{aaai/WangLRRXS20}. 

A quantitative method to evaluate covariate complexity is measuring \textbf{Covariate Homogeneity}. Generally, this implies evaluating how consistently a covariate represents a predefined human-interpretable concept. The exact implementation mainly depends on the type of explanation and type of data. For example, in case of image data, the human-interpretable concepts can be labeled object parts (e.g.~legs, beak and tail of an animal) such that the interpretability of a cluster can be evaluated by checking whether it consistently represents the same part semantics for different images. 
Given an annotated dataset with predefined concepts (such as object parts), the Intersection over Union~\cite{cvpr/ZhangWZ18a, iccv/SunRS19, nips/MuA20, cvpr/FongV18, cvpr/BauZKO017, cvpr/KanehiraTIH19} or distance~\cite{aaai/ZhangCSWZ18} between learned covariates (e.g.~prototypes) and interpretable concepts can be calculated. Zheng et al.~\cite{iclr/ZhengPBH19} evaluate the predictive power of their features for predicting human generated features, and Fyshe et al.~\cite{acl/FysheTMM14} measure the distance from their learned representations to a semantic ground-truth provided by humans. In case of clusters, the purity of a cluster w.r.t. ground-truth labels can be measured~\cite{nips/LakkarajuL16, acl/EskenaziLZ18, icml/ShiZM020, iclr/FortuinHLSR19}.

Homogeneity is also relevant for the explanation type `Disentanglement'. Such explanation methods aim to learn a latent representation that is disentangled such that each dimension corresponds to an interpretable concept, after which these representations can be used in downstream tasks. \textbf{Covariate Homogeneity} evaluates in this case the extent to which each dimension corresponds to exactly one (interpretable) concept, factor or attribute. Examples of such interpretable factors include color of an object in an image, or the presence of a smile on a face~\cite{cvpr/EsserRO20}. This can be quantified by measuring the mutual information between a dimension and a concept~\cite{kdd/GuoZQWSY20}. Ideally, a dimension has high mutual information with a single concept and zero mutual information with all other concepts.
Others vary exactly one dimension and keep all others fixed. Then, the output variance per concept can be measured~\cite{cvpr/EsserRO20}, or the accuracy of a classifier that should predict the index of the specific factor of variation (ceteris paribus)~\cite{kdd/GuoZQWSY20}. Another approach, originally suggested by~\cite{pmlr-v80-kim18b}, is to generate data while keeping exactly one covariate fixed and varying the others, and evaluate whether the variance in one dimension is exactly zero~\cite{cvpr/LiuLZKWBRC20, ijcai/BeyazitTYT020}. 

Another method is a specific metric to evaluate the \textbf{Covariate Regularity} of explanations. Yu and Varshney~\cite{iclr/YuV17} argue that a decision rule is easier to memorize if it is less entropic and therefore measure the Shannon entropy of a rule's feature distribution. Tseng et al.~\cite{nips/TsengSK20} measure the Shannon entropy of feature importance scores in order to indicate how noisy the feature attributions are.

\subsection{Functionally Evaluating Compactness}
\label{sec:eval_compactness}
Many authors have evaluated the compactness of their explanations by measuring their \textbf{Size} (absolute) or sparsity (relative), since explanation should not overwhelm a user. The implemented metric usually depends on the type of data and type of explanation. Examples include: the number of features in an explanation~(e.g.~\cite{aaai/IgnatievNM19, kdd/Ribeiro0G16, nips/PlumbMT18, nips/CrabbeZZS20, aaai/ShakerinG19, nips/Wang18}), average path length in a decision tree~(e.g.~\cite{nips/LageRGKD18, aaai/WuHPZ0D18, aaai/WuPHKCZ0D20}),  reduction w.r.t. complete data sample~(e.g.~\cite{aaai/IgnatievNM19, icml/ChalasaniC00J20, ijcai/AlbiniRBT20}), or the number of decision rules in a set~(e.g.~\cite{kdd/LakkarajuBL16, acl/LiLLLHS20, ijcai/ZhangR19, nips/RawalL20, icml/Wang19, aaai/ShakerinG19}).
Additionally, some evaluate the \textbf{Redundancy} of their explanations. A lower overlap of information within the explanation would signify higher interpretability. Redundancy can be measured with information gain~\cite{kdd/LeW020}, or the overlap ratio~\cite{nips/TsangR020, kdd/LakkarajuBL16}.

In case of counterfactual explanations, compactness can be evaluated by measuring \textbf{Counterfactual Compactness}. A counterfactual explanation, usually an outcome explanation, shows what should change in the input in order to change the corresponding prediction of model $f$, and therefore also addresses Contrastivity. Counterfactual Compactness quantifies how \emph{much} needs to be changed for a different outcome. Generally, as few changes as possible are desired to generate a compact counterfactual explanation. Counterfactual Compactness can be quantified by measuring the distance between the input and counterfactual explanation~\cite{ijcai/KanamoriTKA20, icdm/KarlssonRPG18} or the number of transformations that are required~\cite{icml/GoyalWEBPL19, kdd/LeW020} such as the number of features that need to be changed. Note that a counterfactual explanation with a distance of 0 implies that the contrastivity (and specifically the target sensitivity) is also 0: apparently nothing needs to be changed in the explanation for a different prediction. 

\subsection{Functionally Evaluating Composition}
\label{sec:eval_composition}
The Co-12 property Composition describes the format and organization of an explanation and focuses on \emph{how} something is explained and presented. Composition is often qualitatively discussed by claiming that the predictive model has an interpretable architecture, or by showing anecdotal evidence, for example by concluding that introduced heatmaps are ``the most crisp''~\cite{iclr/KindermansSAMEK18}, or by making specific design choices such as considering the colors used in explanations to make them easier to analyze~\cite{acl/GonenJSG20}. Composition can also be evaluated with users, as shown in Supplementary Material. We identified one functionally-grounded quantitative evaluation method which we term \textbf{Perceptual Realism}. Perceptual Realism for images can be evaluated by using Fréchet Inception Distance (FID) scores~\cite{NIPS2017_8a1d6947_fid} to evaluate the quality of generated images. FID measures the similarity between generated images and real, original images and has been shown to be consistent with human judgment~\cite{NIPS2017_8a1d6947_fid}. FID scores are generally used to evaluate Representation Synthesis explanations generated by GANs~\cite{cvpr/EsserRO20, iclr/SinglaPCB20}. For textual explanations, composition is in most cases indirectly evaluated by measuring standard metrics such as BLEU~\cite{papineni-etal-2002-bleu} and ROUGE~\cite{lin-2004-rouge} with respect to a ground-truth explanation (as incorporated in the `Alignment with Domain Knowledge' evaluation method). However, the perplexity metric as used in e.g.~\cite{nips/CamburuRLB18} does not need a reference input and therefore evaluates the quality of a text in a similar fashion as the FID scores for images. 

\subsection{Functionally Evaluating Confidence}
\label{sec:eval_confidence}
To check whether the explanation contains probability or uncertainty information, authors usually make a design choice whether their XAI method will contain a confidence measure regarding the output of model $f$ or the likelihood of the explanation. Only two papers in our set of included papers explicitly evaluated their confidence information. Schwab and Karlen~\cite{nips/SchwabK19} introduce a feature importance method that produces uncertainty estimates. They assess the \textbf{Confidence Accuracy} of the uncertainty estimates by measuring their correlation with ground-truth changes in outputs when masking features in held-out test samples. As additional baseline, they compare with random uncertainty estimates. Ghalwash et al.~\cite{kdd/GhalwashRO14} evaluate their uncertainty estimates for time series classification by analyzing the correlation between uncertainty thresholds and the model's accuracy, and additionally evaluate how the uncertainty evolves over time. 

\subsection{Functionally Evaluating Context}
\label{sec:eval_context}
The Co-12 property Context takes the user and their needs into account. Naturally, this is evaluated with user studies (as also shown in Supplementary Material). However, we identified two quantitative evaluation methods where user studies are not necessary. The first method is \textbf{Pragmatism}, which quantifies, based on domain knowledge, the degree of difficulty for an individual to act upon the suggestions in a counterfactual explanation~\cite{www/PawelczykBK20, nips/RawalL20}. The underlying intuition is that there might exist various counterfactual explanations that show what should be changed in order to get a different prediction from model $f$. Whereas the Counterfactual Compactness method evaluates the compactness of this counterfactual with respect to the original input, the Pragmatism method also takes the user's context into account. A cost per feature quantifies the degree of difficulty for a user to change that feature. This cost can also be infinite: e.g., a person cannot get younger so a counterfactual explanation showing that the user should decrease their age is not actionable and hence not pragmatic. In addition to the cost per feature, also the degree or cost of a feature change could be taken into account, related to Compactness. Rawal and Lakkaraju~\cite{nips/RawalL20} explain that it is probably easier for a user to increase income by 5K than 50K. 

The second method involves \textbf{Simulated User Studies}. Ribeiro et al.~\cite{kdd/Ribeiro0G16} perform two experiments to evaluate whether users would 1) trust predictions and 2) be able to use explanations as model selection method. In their first experiment, they define some input features as being ``untrustworthy'' and assume that end users do not want such features to be used by the predictive model. They evaluate whether the prediction changes when all untrustworthy features are removed from the explanation. Their second experiment trains two predictive models that have a similar validation accuracy but one performs worse on the test set. They evaluate whether their explanation method can be used to identify the better model by revealing artificially introduced spurious correlations. Singla et al.~\cite{iclr/SinglaPCB20} evaluate in a similar manner the performance of their explanation method in identifying biases in data.

\subsection{Functionally Evaluating Coherence}
\label{sec:eval_coherence}
To quantitatively evaluate whether explanations generated by an XAI method align with domain knowledge, general beliefs and consensus, they are often compared with some expected explanations framed as `ground-truth', which we call \textbf{Alignment with Domain Knowledge}. This ground-truth is contained in an annotated dataset. 
For imaging data, often `location coherence' is evaluated by comparing a heatmap or localization explanation with ground-truth object bounding boxes, segmentation masks, landmarks or human attention maps. Correspondence between the ground-truth and the explanation can then be quantified with the Intersection over Union~(also known as Jaccard index, e.g.~\cite{aaai/NamGCWL20, cvpr/WangV20, cvpr/ZengLSSYCU19, iccv/FongV17, iclr/ChangCGD19, kdd/DuLSH18, cvpr/BauZKO017}), outside-inside relevance ratio~(e.g.~\cite{aaai/NamGCWL20}), point localization error~(e.g.~\cite{cvpr/HuangL20, cvpr/JakabGBV20, cvpr/ZengLSSYCU19}), pointing game accuracy (whether a point falls into a ground-truth region, e.g.~\cite{kdd/DuLSH18, cvpr/HuangL20}) or rank correlation with human attention maps~\cite{aaai/PatroAN20, iccv/PatroLPN19}. For textual explanations, standard natural language generation metrics such as ROUGE~\cite{lin-2004-rouge} and BLEU~\cite{papineni-etal-2002-bleu} are often used to evaluate the overlap of the generated explanations with a ground-truth text, e.g.~\cite{acl/AtanasovaSLA20, acl/LiuYW19, acl/RajaniMXS19, cvpr/ChuangL0F18, aaai/WickramanayakeH19, cvpr/ParkHARSDR18, ijcai/ChenW0PSAC20, ijcai/ChenW0WBWC19, ijcai/LeL20, nips/CamburuRLB18, www/SunWZFHW20, cvpr/DongSZZ17}. Indirectly, these metrics also evaluate composition since it has been shown that ROUGE and BLEU correlate with human judgments on fluency of a text~\cite{fabbri_2021_eval_summarization}.
For real-valued explanations such as feature importance, one could measure (rank) correlation between the generated explanation and ground-truth annotation in the dataset~(e.g.~\cite{icdm/JhaWXZ18, iccv/PatroLPN19, iccv/SelvarajuLSJGHB19, iclr/JinWDXR20, icml/AndersPDMK20, nips/TsangR020, nips/TsengSK20, acl/ShahbaziFGT20, aaai/PatroAN20, icml/AnconaOG19, iclr/JinWDXR20, acl/MohankumarNNKSR20, nips/BassSSTSR20}). We would like to highlight that some common correlation metrics were criticized, such as Kendall's tau which could be misleading at the tail end of distributions~\cite{acl/MohankumarNNKSR20} and Spearman correlation has limitations for global rankings~\cite{acl/GonenJSG20}.

As alternative to evaluating the Alignment with Domain Knowledge, for example when a ground-truth is not available, one could evaluate coherence by calculating the \textbf{XAI Methods Agreement}.
Existing XAI methods, which are already established within the community and/or proven to adhere to certain desirable properties, are then usually considered as ground-truth, and those explanations are compared with the explanations from other methods. 

For both \textbf{Alignment with Domain Knowledge} and \textbf{XAI Methods Agreement}, we would like to emphasize that these methods only evaluate Coherence with respect to expectations, and not Correctness with respect to the predictive model $f$. Hence, a coherent explanation could be incorrect and vice versa. For example, an explanation highlighting snow in the background to distinguish between a husky and a wolf (example from Ribeiro et al.~\cite{kdd/Ribeiro0G16}), would score low on location coherence w.r.t. object segmentation masks, but is correctly showing the reasoning of this bad classifier. It is therefore good practice to evaluate multiple Co-12 properties and specifically evaluate correctness and coherence independently, as also discussed in Section~\ref{sec:plausibility_vs_correctness}.

\subsection{Functionally Evaluating Controllability}
\label{sec:eval_controllability}
Controllability addresses the interactivity of explanations, which is applicable to e.g. conversational explanation methods~\cite{ijcai/ChenW0PSAC20, acl/MoonSKS19}, interactive interfaces~\cite{aaai/RustamovK18, nips/TsangR020}, human-in-the-loop explanation learning methods (e.g.~as~\cite{cvpr/DongSZZ17}) or methods that enable the user to correct explanations~\cite{www/Wang0FNC18}. We note that XAI methods can also have (hyper)parameters to tune the explanations, such as a regularizer for explanation size, but we do not consider such parameters as being evaluation methods for Controllability. The evaluation of Controllability is usually qualitative by discussing why the controllable format improves the quality of the explanations, or by only showing an example of the Controllability. We identified two papers that \emph{quantified} Controllability by measuring the improvement of explanation quality after human feedback: \textbf{Human Feedback Impact}. Chen et al.~\cite{ijcai/ChenW0PSAC20} and Dong et al.~\cite{cvpr/DongSZZ17} measure the accuracy of their textual explanations after iterative user feedback. Although users are involved in this evaluation method, it is not a standard user study since the user is seen as a system component: the XAI methods use optimization criteria that require humans-in-the-loop for optimal output. 
Additionally, Chen et al.~\cite{ijcai/ChenW0PSAC20} define the ``Concept-level feedback Satisfaction Ratio'' which measures whether concepts for which a user has indicated to be interested in are present in the explanation, and whether concepts where the user is not interested in are removed from the explanation. This Satisfaction Ratio does not require direct user feedback, and could also be applied to an existing dataset with user interests.

\section{Implications and Research Opportunities}
\label{sec:research_opportunities}
We strongly believe that explainable AI has great potential: XAI can \emph{justify} algorithmic decisions, XAI can allow users to \emph{control} and \emph{improve} systems by identifying and correcting errors, and XAI can contribute to \emph{knowledge discovery} by revealing learned patterns~\cite{adadi_peeking_2018}. However, to reach this full potential, \textbf{XAI methods should be extensively validated} in order to ensure that they are reliable and useful. Our analysis has shown that the field has been maturing the last few years, but at the same time we also see that explainability is still often presented as a binary property. We argue that explainability is a multi-faceted concept and make this explicit with \textbf{our Co-12 properties} describing different conceptual aspects of explanation quality. Instead of evaluating only one property, it is essential to get insight into, and preferably quantify, all properties such that an informed trade-off can be made. ``Best is not directly a judgment of truth but instead a summary judgment of accessible explanatory virtues''~\cite{josephson1996abductive}. 
In practice, such a multi-dimensional overview could be implemented as a radar chart or as a set of \emph{consumer labels} as proposed by Seifert et al.~\cite{seifert_towards_2019} that comprehensively and concisely conveys the strengths and weaknesses of the explanation or explanation method. Our \textbf{collection of identified evaluation methods} also shows that quantitative evaluation methods exist for each of the Co-12 properties. On the other hand, our analysis reveals that the majority of XAI evaluation focused on evaluating Coherence, Completeness, Compactness or Correctness. We hope that our collection of evaluation methods will stimulate and \textbf{facilitate a more complete and inclusive evaluation} in order to objectively validate and compare new and existing XAI methods. Eventually, we are convinced that XAI methods should be kept to minimal standards, similarly as such standards exist for predictive models. Our overview of evaluation methods provides researchers and practitioners with concrete tools to evaluate every Co-12 property while using unified terminology, and therefore contributes to standardization. 
We also see a research opportunity to develop new evaluation methods for Co-12 properties that are currently insufficiently addressed, and to develop variants on existing evaluation methods to make them suited for different types of data and explanations. 

Besides, we acknowledge that it might be unreasonable to expect an XAI method to score well on all Co-12 properties. In practice,
\textbf{trade-offs between desired explanation properties} will have to be made when developing an XAI method. Coherence might contradict with Correctness, as discussed in Section~\ref{sec:plausibility_vs_correctness}, and Completeness and Compactness might be considered diametrical opposites. The application domain or practical feasibility can determine which Co-12 properties should be emphasized.
Herman~\cite{herman_promise_2017} proposes to optimize explanations for content-related properties first (correctness and completeness in particular), without making effort to simplify the explanation. ``This separation of concern encourages more rapid innovation and reduces the cost of evaluation''~\cite{herman_promise_2017}. Subsequently, a second step can consist of altering the explanation to ``incorporate human cognitive function, user preferences, and expertise into the explanation''~\cite{herman_promise_2017}. Eventually, any trade-off can be made as long as it is sufficiently motivated. Insights from \textbf{other research areas}, such as social sciences, psychology and HCI, can also provide the XAI community with more guidance regarding what aspects of an explanation are important to evaluate. Combining strengths in multi-disciplinary collaborations can subsequently result in innovative XAI evaluation methods.

Additionally, we think that our collected set of evaluation methods can not only be used for thorough evaluation, but also for multi-dimensional \textbf{optimization of interpretability}. 
Some papers already optimize for interpretability by using a regularization term or objective function during training of the predictive model (e.g.~\cite{aaai/WuHPZ0D18, aaai/WuPHKCZ0D20, icml/ChalasaniC00J20, kdd/Tao0WFYZ019, nips/PlumbACPXT20, ijcai/RossHD17, iclr/YuV17, kdd/LakkarajuBL16}) or by relating rewards to explainability and presentation quality in reinforcement learning~(e.g.~\cite{icdm/WangCYWW018}). However, in practice these interpretability optimizers usually involve only one to two Co-12 properties.
Interestingly, some interpretability regularizers are not (yet) used as evaluation metric although their quantitative nature would make them suited to be used as XAI evaluation method as well. For example, Park et al.~\cite{icml/ParkCZYY20} optimize for interpretability with spatial auto-correlation, which we have not seen as evaluation metric but might be suited for evaluating heatmaps. This shows that the optimization and quantitative evaluation of explanation methods are closely related. We recognize a research opportunity to study how evaluation methods can be incorporated in the training process of predictive models, in order to tune the so-called ``accuracy-interpretability trade-off'' \emph{during} training instead of only analyzing it afterwards.  
Also the quantitative evaluation methods where user involvement is required can be used for optimizing an interpretable model by adopting a human-in-the-loop approach~(as done in e.g.~\cite{ijcai/ChenW0PSAC20, nips/LageRGKD18}).

Lastly, we believe that our annotated dataset containing the categorization of 312 XAI papers (such as type of data and explanation, as shown in Fig.~\ref{fig:review_protocol_categorisation}) is a rich source of information and can be a useful starting point for more in-depth research. \textbf{Our dataset is therefore publicly available} at \url{https://utwente-dmb.github.io/xai-papers/}, such that others can efficiently collect XAI papers that adhere to specific criteria such that subtopics can be analyzed in more detail. 

\paragraph{Acknowledgments}
We would like to thank Ziekenhuis Groep Twente (ZGT) for supporting this project, and Marion Koelle for the useful suggestions to improve the structure of this work.

\clearpage

\section*{Supplementary Material}
\label{sec:supplementary}
Supplementary material presenting quantitative evaluation methods with user studies (Table~\ref{tab:evaluation_methods_userstudies}), providing a more detailed discussion on the extent and nature of XAI research, and describing our paper collection and reviewing process in more detail. 

\subsection{Quantitative Evaluation with User Studies}
\label{sec:evaluation_methods_humans}
\renewcommand{\arraystretch}{1.0}
\afterpage{
\addtolength{\tabcolsep}{-3pt} 
\small{
\begin{longtable}[H]{p{0.69\linewidth}@{\hskip 5pt}|@{\hskip 2pt}*{6}{cg}}
    \caption{Descriptions of quantitative evaluation methods with user studies, with references to papers that apply this method. Bold check mark indicates prominent Co-12 property.}\\
    \label{tab:evaluation_methods_userstudies}
        \textbf{Name and Description of Quantitative Metric, with References} &\rot{Correctness} &\rot{Output-completeness} &\rot{Consistency} & \rot{Continuity} &\rot{Contrastivity} & \rot{Covariate Complexity} & \rot{Compactness} &\rot{Composition} & \rot{Confidence} & \rot{Context} & \rot{Coherence} & \rot{Controllability}\\
        \toprule
        \endhead
        \textbf{Forward Simulatability} \newline Given an explanation (and possibly the corresponding input sample), ask users to guess or identify the model's prediction~(\textit{human-output-completeness}). Additionally, the user's prediction speed can be measured, or the difference in simulation accuracy between whether or not explanations are shown.  \footnotesize{\cite{aaai/AkulaWZ20, aaai/KimPRS15, aaai/Madumal0SV20, acl/ChenZJ20, acl/HaseB20, acl/RajaniMXS19, iclr/ChenSWJ19, kdd/LiangBCBW20, kdd/MingXQR19, nips/KimKK16, nips/LageRGKD18, nips/PalejaSCG20, nips/Wang18, nips/WangN19, ijcai/HoernleGGLRR20, aaai/Ribeiro0G18, acl/AtanasovaSLA20, icml/ChenSWJ18, nips/RamamurthyVZD20, kdd/LakkarajuBL16, www/WuWYLZ18, ijcai/0002DSJNICFB19, iclr/PuriVGKDK020, sigir/RamosE20}} && \boldcheckmark &  &\checkmark&&&\checkmark&\checkmark&& \checkmark &&\\
        \textbf{Teaching Ability} \newline Train users with explanations to understand the model's reasoning, after which humans should predict the ground-truth for a new data instance \emph{without} having an explanation. Additionally, the user's prediction speed can be measured. \footnotesize{\cite{icml/GoyalWEBPL19, cvpr/WangV20}} &  \phantom{\boldcheckmark}& \checkmark &\phantom{\boldcheckmark} &\checkmark&\phantom{\boldcheckmark}&\phantom{\boldcheckmark}&\phantom{\boldcheckmark}&\phantom{\boldcheckmark}&\phantom{\boldcheckmark}&\boldcheckmark&\phantom{\boldcheckmark}&\phantom{\boldcheckmark}\\
        \textbf{Subjective Satisfaction} \newline Ask users to rate explanations on properties such as satisfaction, reasonableness, usefulness, fluency, relevance, sufficiency and trust. \footnotesize{\cite{aaai/AkulaWZ20, aaai/GaoWW019, aaai/KimPRS15, aaai/Madumal0SV20, acl/PruthiGDNL20, iccv/SelvarajuCDVPB17, iccv/SelvarajuLSJGHB19, icdm/WangCYWW018, iclr/JinWDXR20, ijcai/ChenW0PSAC20, ijcai/ChenW0WBWC19, ijcai/LeL20, kdd/TolomeiSHL17, kdd/ZhaoGS19, nips/AdebayoMLK20, sigir/MiJ19, sigir/RamosE20, sigir/WangWJY18, kdd/Ribeiro0G16, nips/VuT20, nips/ZhouHZLSXT20, acl/HaseB20, nips/PalejaSCG20, sigir/TaoJWW19, nips/DhurandharCLTTS18, cvpr/BauZKO017, acl/KumarT20}} & & & &\phantom{\boldcheckmark}& & \checkmark & \checkmark&\checkmark&&\checkmark&\boldcheckmark&\\ 
        \textbf{Subjective Comparison} \newline Show users explanations from different XAI methods (or explanations from humans) and evaluate which method is perceived as being better (in terms of e.g. perceived accuracy, usefulness or understandability). \footnotesize{\cite{acl/AtanasovaSLA20, aaai/ChenZQ19, acl/MohankumarNNKSR20, cvpr/WangS0H18, kdd/ShuCW0L19, sigir/TaoJWW19, www/ChenZLM18, acl/LiuYW19, nips/VuT20, cvpr/ParkHARSDR18, nips/GhorbaniWZK19, nips/JeyakumarNCGS20}} & & & & & & & \checkmark&\checkmark&&\checkmark&\boldcheckmark&\\ 
        \textbf{Perceived Homogeneity} \newline Ask users to evaluate the purity or disentanglement of explanations, by e.g. verifying that a dimension corresponds to a single interpretable factor. \footnotesize{\cite{acl/EskenaziLZ18, icml/ShiZM020, icml/VoynovB20, aaai/ZhangCSWZ18}} & &  & & & & \boldcheckmark & &&&&\checkmark&\\
        \textbf{Intruder Detection} \newline Given an explanatory prototype or disentangled concept, show users a set of instances of which one is an intruder, and ask which instance does not correspond with the explanation. \footnotesize{\cite{nips/GhorbaniWZK19, aaai/SubramanianPJBH18, acl/PanigrahiSB19}} & & & & & &\boldcheckmark& &&&&\checkmark&\\ 
        \textbf{Synthetic Artifact Rediscovery} \newline A controlled experiment where a property of the predictive model is changed, after which it is evaluated whether humans can reveal this property with the help of explanations. \footnotesize{\cite{acl/SydorovaPR19, nips/RawalL20, iclr/SinghMY19, kdd/Ribeiro0G16}} & & & & & & & &&&\checkmark&\boldcheckmark&\\ 
        \bottomrule
\end{longtable}}
\addtolength{\tabcolsep}{3pt}
}
\normalsize

Whereas the main paper focuses on automated quantitative evaluation methods, Table~\ref{tab:evaluation_methods_userstudies} summarizes the main quantitative evaluation methods we identified that were applied in user studies. Generally, we can distinguish between \emph{subjective} evaluation and \emph{objective} evaluation. The subjective methods usually evaluate Coherence by measuring how a user perceives an explanation. In contrast, the Forward Simulatability, Teaching Ability, Intruder Detection and Synthetic Artifact Rediscovery are objective evaluation methods. The Forward Simulatability method is comparable to the functionally-grounded Preservation Check: instead of evaluating whether the predictive model gives the right output given the explanation, the user acts as a surrogate model to evaluate whether the explanation is output-complete for a user. The Teaching Ability is a related approach but also evaluates whether the explanations are generalizable such that a user can, after being trained with explanations, make a correct prediction \emph{without} having an explanation. We refer to other work (e.g.~\cite{lage_evaluation_2019, hoffman_metrics_2019,chromik_taxonomy_2020}) for a more detailed discussion on evaluation with human subjects. 

\subsection{Categorization and Analysis of XAI Methods}
\label{sec:sub_results_g2}

Figure~\ref{fig:barcharts_all_G2} visualizes the categorization of papers per dimension. Generally, it can be seen that the top-3 in each dimension covers the majority of the papers. Especially the imbalance regarding the types of models to be explained is striking, since a large majority of the literature (75\%) focuses on explaining neural networks, which could be due to their black box nature and state-of-the-art performance. The `other' category follows with 17\%, which usually involves models which are introduced in the paper and which are specifically designed to explain a certain prediction task. 15\% of the papers are XAI methods which can, according to the authors, be applied to any predictive model. This model-agnostic category includes for example methods that explain a latent representation which can come from any model. 

Our analysis also shows that there is a high diversity in explanation types. Interesting is that the top-3 explanation types are dominated by feature importance methods: 27\% of the papers use standard feature importance scores, followed by heatmaps (2-dimensional feature importance) and localization (binary feature importance).
The types of data that are given as input to predictive model $f$ are also diverse, although images, text and tabular data are most used. Additionally, we found that image data make up the majority for heatmap explanations whereas textual input is mostly used for textual explanations.
Feature importance seems to be the most generally applicable explanation type since it is used for all data types, and is also often used for `Other' data types that do not fall into the predefined categories.

We also categorized the main task of the predictive models. These statistics might be highly influenced by our selection of publication venues and also the difficulty of a task can play a role. A great majority (63\%) of the papers presents a model for classification. Besides the fact that classification is a broad concept, it can also be related with the popularity of outcome explanations: explaining a particular classification decision can be a good use case for outcome explanations. 
As shown in Figure~\ref{fig:barcharts_all_G2}, the type of problem that is addressed in 64\% of the papers is the outcome explanation, meaning that an explanation aims to explain a single prediction. In contrast, the second most addressed type, model inspection (30\% of the papers), gives global explanations about some property of predictive model $f$, such as global feature importance. This can imply that users or model developers are more interested in understanding specific decisions than getting global insights, or that global XAI methods are more difficult to develop. The transparent box designs are by themselves already interpretable, such as a decision tree. Few papers present a model explanation, meaning that a second, interpretable model is learned to mimic the output of the black box. We hypothesize that the model explanation is not often addressed since there is no guarantee that the original black box and the interpretable surrogate model agree on their internal reasoning~\cite{rudin2019stop}. Hence, the resulting explanation could be incorrect with respect to the workings of the original predictive model. Interesting to add is that the outcome explanation is also the dominant problem for papers that do not introduce, but apply or evaluate an existing XAI method. Specifically, 82\% of the papers excluded by the filter focus on outcome explanations. This might indicate that outcome explanations are easier to apply and compare among different XAI methods than global explanations.

Lastly, we see that the type of method used to explain is dominated by two types: post-hoc explanation methods that aim to explain an already trained model, and interpretability built into the predictive model. Built-in interpretability is a broad category ranging from models that are intrinsically interpretable to additions or restrictions to the predictive model architecture. The latter includes attention mechanisms, regularizers in the training process to improve interpretability, or a combined architecture that merges the prediction and explanation task. A minority produces explanations based on supervised explanation training, such as optimizing the XAI method to generate explanations that are similar to ground-truth explanations from a dataset. Interesting to note is that all supervised explanation training methods in our dataset produce outcome explanations. 

\subsubsection{XAI Evaluation Practice per Venue}
\begin{figure}[!tb]
    \centering
    \includegraphics[width=0.95\linewidth]{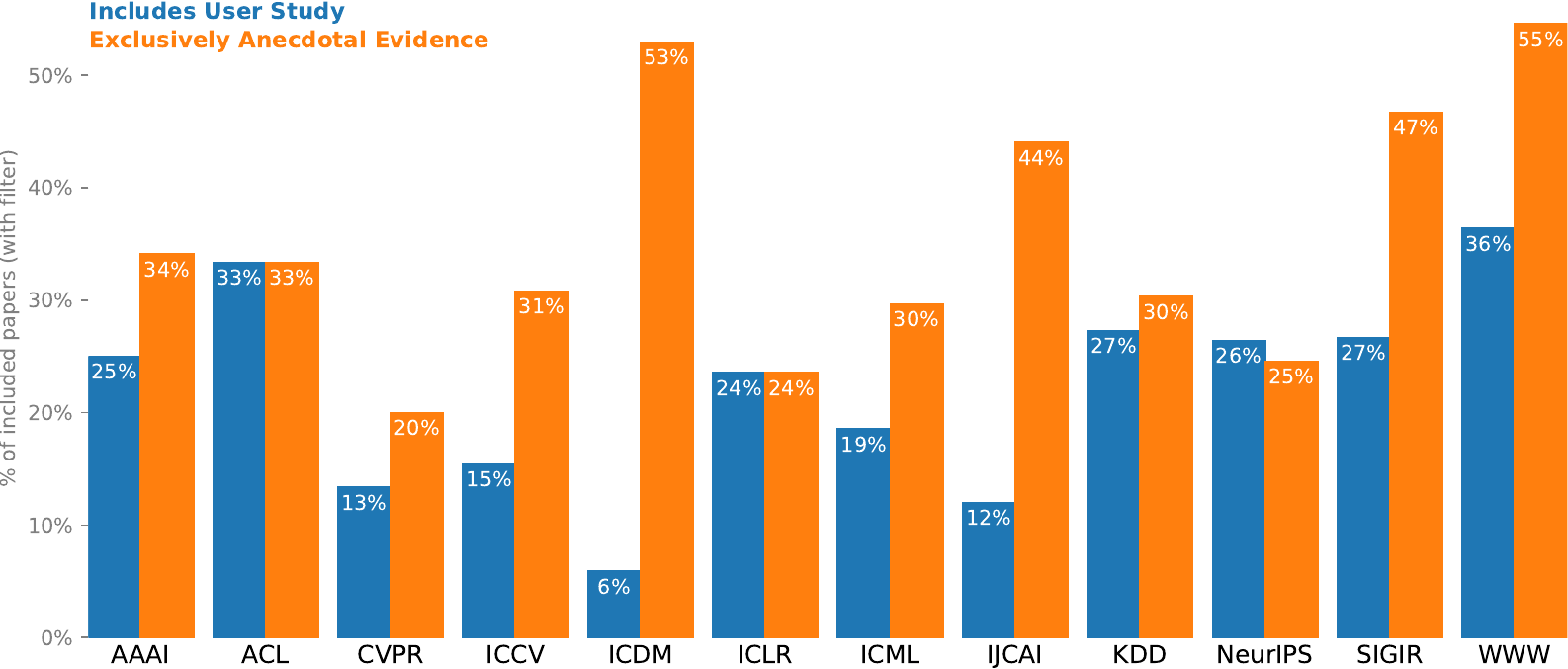}
    \caption{Fraction of papers per venue that introduce an XAI method and evaluate with user studies (blue) or exclusively with anecdotal evidence (orange). The percentages are based on the total number of included papers per venue that introduce an XAI method.}
    \label{fig:stacked_barchart_pervenue}
\end{figure}

Figure~\ref{fig:stacked_barchart_pervenue} provides more insight into the XAI evaluation practices per venue. It shows that generally the more application-oriented conferences evaluate only with anecdotal evidence in roughly half of the cases. In contrast, the more theoretical venues have percentages around 20\%-30\%. We could not see a clear trend regarding the usage of user studies, although the relative differences between conferences is striking. 

\subsection{Methodology}
\label{sec:methodology_suppl}
This section describes in more detail our paper collection process, inclusion and exclusion criteria, and reviewing process.

\subsubsection{Identification of Paper Candidates}
\label{sec:strategy_initial_selection}
We collected papers in a structured manner to provide both quantitative and qualitative insights about the XAI domain based on a large corpus of scientific work on XAI evaluation methods. 
Since the literature on XAI is highly diverse and distributed across different (sub)disciplines, we selected literature by filtering on publication venue, date and title.

\textbf{Publication Venue.} To obtain a representative and sufficiently large, yet feasible selection of papers, we considered literature from all areas of AI, ranging from Computer Vision and Information Retrieval to Natural Language Processing and Data Mining. Specifically, we considered literature from the following twelve prominent\footnote{As indicated by their A* ranking according to the CORE 2021 rankings portal (\url{http://portal.core.edu.au/conf-ranks/}).} conferences: AAAI, IJCAI, NeurIPS (formerly NIPS), ICML, ICLR, CVPR, ICCV, ACL, WWW, ICDM, SIGKDD (also called KDD), SIGIR. 
This selection criterion also implies that all work included in the set is original, peer-reviewed and written in English. We are aware of the fact that we exclude relevant papers published at other venues, but do believe that our selection of venues is sufficiently representative to enable extrapolation of our results and conclusions to XAI literature as a whole.

\textbf{Publication Year.} We scoped our selection to work published from  2014 to 2020. This criterion is motivated by the fact that XAI has gained renewed interest since the emergence of deep learning, and the fact that annual international conference series dedicated exclusively to explainability or interpretability were organized from 2014 onwards~\cite{adadi_peeking_2018}.

\textbf{Keywords in Title.} We conducted a keyword search in publication titles with the following search query: \texttt{explain* OR explanat* OR interpret*} to capture terms including \emph{explainable}, \emph{explaining}, \emph{explanation}, \emph{interpretable} and \emph{interpretability}.
We are aware of the fact that this query excludes papers with related terms (such as \emph{intelligibility} and \emph{transparency}), and papers that specify a specific explanation method (such as \emph{feature attribution}). However, to reflect time and resource constraints, we aimed for high precision instead of high recall. For a similar reason, we did not consider keywords in abstracts or full-texts, since we found that searching for general terms as \emph{explain} leads to a surge in irrelevant results. 

\textbf{Final Search Query.} We used the search engine of computer science bibliography DBLP\footnote{\url{https://dblp.org/}} to collect the initial selection of papers. Combining the aforementioned criteria results in the following query to DBLP:\\
\texttt{\small{explain | explanat | interpret year:2020: | year:2019: | year:2018: | year:2017: | year:2016: | year:2015: | year:2014: venue:ICDM: | venue:KDD: | venue:NIPS: | venue:NeurIPS: | venue:CVPR: | venue:ICCV: | venue:AAAI: | venue:IJCAI: | venue:SIGIR: | venue:ACL: | venue:WWW: | venue:ICLR: | venue:ICML:}}\\
This search, conducted on 4th of May 2021, resulted in 606 papers.

\subsubsection{Inclusion and Exclusion}
\label{sec:inclusion_exclusion}
Before screening the main content of each paper according to inclusion criteria, we applied an exclusion criterion since we found that the search result from the query contained more than the main conference papers. 

\textbf{Exclusion.} We manually excluded companion papers, which include extended abstracts and papers from workshops, doctoral consortium and early career tracks, invited talks, senior member presentations, demonstrations, companion proceedings, challenges and tutorials. Applying this exclusion criterion to the initial query result resulted in 494 papers. 

\textbf{Inclusion.} The resulting 494 papers are screened according to an inclusion criterion, in order to only include relevant papers in our analysis. With our inclusion criterion, we focus on papers in the explainable AI domain and therefore exclude papers that use the terms ``explain'' or ``interpret'' in other contexts.

\begin{inclusion_criterion*}
Original work introducing, applying and/or evaluating one or more methods for explaining a machine learning model.
\label{inclusion_criterion_frame}
\end{inclusion_criterion*}
With ``introducing'', we mean that the work presents a new method for explaining a machine learning model. The term ``machine learning'' implies learning from data. Since we require that this machine learning model should be explained (see the main paper for our definition of `explanation'), we do not include papers that only explain the \emph{data} rather than explaining how a predictive model does something. 

Applying the inclusion criteria to the set of 494 papers, led to 361 papers being included. Subsequently, we can apply a \textbf{filter} that only selects the papers that \emph{introduce} an XAI method, resulting in 312 papers. We apply this filter to analyze how introduced XAI methods are evaluated when they are first presented. For collecting all evaluation methods, we review all 361 included papers since 49 papers do not introduce a new XAI method, but could contain relevant evaluation metrics to compare and evaluate existing XAI methods. We do not want such papers to skew our quantitative results on XAI methods (Figure~\ref{fig:barcharts_all_G2}), but include them in our evaluation overview for completeness. 

\subsubsection{Inclusion and Reviewing Process.} 
\label{sec:includsion_and_reviewing_process}
Screening papers for inclusion and reviewing them according to a review protocol was a collaborative task. All authors have a background in machine learning and have explainable AI as research interest. Each author reviewed papers individually but there was frequent communication within the team to align and verify inclusion and categorization decisions. 
For 81 papers, the inclusion criteria were checked by two reviewers in order to measure the inter-rater agreement for quality assurance. Besides a random sample of papers that were reviewed twice, the majority in this set were papers where the initial reviewer indicated that they were not confident about the decision, after which another reviewer checked the paper. 
We therefore emphasize that the following agreement metrics can be biased towards lower scores due to the perceived difficulty of the papers in the specific subset. The two reviewers were in agreement on the inclusion decision for 68 out of 81 papers, resulting in a Cohen's kappa $\kappa = 0.625$ and a Matthews correlation coefficient $MCC=0.647$ (the latter is said to be better suited for binary and imbalanced data~\cite{mcc_chicco}). These results indicate substantial agreement~\cite{landis1977measurement}. In case of disagreement or low confidence by both reviewers, discussion took place to come to a final inclusion decision. In addition to disagreement, also more informal discussion took place whenever a reviewer was in doubt on aspects of the review protocol. After reviewing, the first author did an extra check regarding the categorization of evaluation methods of all included papers. 

\clearpage
\bibliographystyle{acm}
\bibliography{allreferences}
\end{document}